\documentclass[letterpaper]{article} %
\usepackage[dvipsnames,table,xcdraw]{xcolor}
\usepackage{aaai2026}  %
\usepackage{times}  %
\usepackage{helvet}  %
\usepackage{courier}  %
\usepackage[hyphens]{url}  %
\usepackage{graphicx} %
\usepackage{natbib}  %
\usepackage{caption} %
\usepackage{algorithm}
\usepackage{algorithmic}

\usepackage{newfloat}
\usepackage{listings}
\DeclareCaptionStyle{ruled}{labelfont=normalfont,labelsep=colon,strut=off} %
\floatstyle{ruled}
\newfloat{listing}{tb}{lst}{}
\floatname{listing}{Listing}
\usepackage{amsmath}
\usepackage{amssymb}
\usepackage{bm}

\usepackage{subcaption}  %
\usepackage{multirow}
\usepackage{booktabs}   %
\usepackage{makecell}   %
\usepackage{changepage} %
\usepackage{comment}

\usepackage{footmisc}  %

\usepackage{pifont}
\newcommand{\cmark}{\ding{51}}
\newcommand{\xmark}{\ding{55}}

\usepackage{tabularx}
\newcolumntype{C}{>{\centering\arraybackslash}X}
\newcolumntype{R}{>{\raggedleft\arraybackslash}X}
\newcolumntype{L}{>{\raggedright\arraybackslash}X}

\makeatletter
\usepackage{xspace}
\def\@onedot{\ifx\@let@token.\else.\null\fi\xspace}
\DeclareRobustCommand\onedot{\futurelet\@let@token\@onedot}

\def\eg{\emph{e.g}\onedot} \def\Eg{\emph{E.g}\onedot}
\def\ie{\emph{i.e}\onedot} 
 
 \def\vs{\emph{vs}\onedot}

\newcommand{\figref}[1]{Fig\onedot~\ref{#1}}
\newcommand{\equref}[1]{Eq\onedot~\eqref{#1}}
\newcommand{\secref}[1]{Sec\onedot~\ref{#1}}
\newcommand{\tabref}[1]{Tab\onedot~\ref{#1}}

\usepackage{csquotes}

\usepackage{multirow}
\usepackage{listings}

\usepackage{adjustbox}

\usepackage{fancybox, graphicx}

\usepackage{color, colortbl}
\definecolor{Gray}{gray}{0.9}
\newcolumntype{g}{>{\columncolor{Gray}}r}
\definecolor{highlight}{rgb}{0.9, 0.9, 1}

\definecolor{tabyellow}{rgb}{1,1, 0.6}
\definecolor{tablightyellow}{rgb}{1,1, 0.8}
\definecolor{taborange}{rgb}{1, 0.8, 0.6}
\definecolor{tabred}{rgb}{1, 0.6, 0.6}
\definecolor{tabhighlight}{rgb}{0.9, 0.9, 1}

\definecolor{light-gray}{gray}{0.7}

\newcommand*{\belowrulesepcolor}[1]{%
  \noalign{%
    \kern-\belowrulesep 
    \begingroup 
      \color{#1}%
      \hrule height\belowrulesep 
    \endgroup 
  }%
} 
\newcommand*{\aboverulesepcolor}[1]{%
  \noalign{%
    \begingroup 
      \color{#1}%
      \hrule height\aboverulesep 
    \endgroup 
    \kern-\aboverulesep 
  }%
} 

\newcommand{\beginsupplement}{
    \setcounter{table}{0}
    \renewcommand{\thetable}{S\arabic{table}}%
    \setcounter{figure}{0}
    \renewcommand{\thefigure}{S\arabic{figure}}%
    \setcounter{equation}{0}
    \renewcommand{\theequation}{S\arabic{equation}}
}

\newcommand{\data}{\mathbf{x}}
\newcommand{\sample}{\mathbf{z}}
\newcommand{\cond}{c}
\newcommand{\noise}{\bm{\epsilon}}

\newcommand{\set}{\mathcal{X}}
\newcommand{\setreal}{\mathcal{X}_\text{real}}

\newcommand{\cgguide}{classifier guidance}
\newcommand{\Cgguide}{Classifier guidance}

\newcommand{\cfgguide}{classifier-free guidance}
\newcommand{\Cfgguide}{Classifier-free guidance}
\newcommand{\cfgabb}{CFG}

\definecolor{tabyellow}{rgb}{1,1, 0.6}
\definecolor{tablightyellow}{rgb}{1,1, 0.8}
\definecolor{taborange}{rgb}{1, 0.8, 0.6}
\definecolor{tabred}{rgb}{1, 0.6, 0.6}

\newcommand\rurl[1]{%
  \href{https://#1}{\nolinkurl{#1}}%
}

\definecolor{cvprblue}{rgb}{0.21,0.49,0.74}

\usepackage[pagebackref,breaklinks,colorlinks,citecolor=ForestGreen]{hyperref}

\newcommand{\ourtitle}{Studying Classifier(-Free) Guidance From A Classifier-Centric Perspective}

\title{\ourtitle}

\author{
    Xiaoming Zhao, Alexander G. Schwing
}
\affiliations{
    University of Illinois Urbana-Champaign \\
    \{xz23, aschwing\}@illinois.edu
}

\usepackage{bibentry}

\begin{document}

\maketitle

\begin{abstract}
\Cfgguide~has become a staple for conditional generation with denoising diffusion models.
However, a comprehensive understanding of \cfgguide~is still missing.
In this work, we carry out an empirical study to provide a fresh perspective on \cfgguide.
Concretely, instead of solely focusing on \cfgguide, we trace back to the root,~\ie,~\cgguide, pinpoint the key assumption for the derivation, and conduct a systematic study to understand the role of the classifier.
On 1D data, we find that both \cgguide~and \cfgguide~achieve  conditional generation by pushing the denoising diffusion trajectories away from decision boundaries, \ie, areas where conditional information  is usually entangled and is hard to learn.
To validate this classifier-centric perspective on high-dimensional data, we assess whether a flow-matching postprocessing step that is designed to narrow the gap between a pre-trained diffusion model's learned distribution and the real data distribution, especially near decision boundaries, can improve the performance.
Experiments on various datasets verify our classifier-centric understanding.
\end{abstract}

\section{Introduction}

Conditional generation,~\eg, class-to-image, text-to-image, or image-to-video, is omnipresent as it provides a compelling way to control the output. %
Ideally,  conditional generation results are both \textit{diverse} and of \textit{high-fidelity}. Namely, the generative models' outputs  align with the conditioning information perfectly and diligently  follow the training data diversity.
However, there is a trade-off between high-fidelity  and diversity: without constraining diversity  there are always possibilities to sample from areas on the data distribution manifold that are not well-trained.
Thus, trading diversity for fidelity is a long-standing problem and the community has developed various approaches,~\eg,~the truncation trick for generative adversarial nets (GANs)~\cite{Brock2018LargeSG,Karras2018ASG}, low-temperature sampling for probabilistic models~\cite{Ackley1985ALA}, or temperature control in large language models~\cite{achiam2023gpt,dubey2024llama}.

More recently, to trade  diversity and  fidelity in  denoising diffusion  models~\cite{song2019gen,vincent2011connection,Ho2020DenoisingDP,Kingma2021VariationalDM},  several techniques have been developed~\cite{dhariwal2021diffusion,Hong2022ImprovingSQ,Kim2022RefiningGP, Dinh2023RethinkingCD, Dinh2023PixelAsParamAG}, from which classifier-free guidance~\cite{ho2022classifier} emerged as the de-facto standard. 
\Eg,~\cfgguide, especially at sufficient scale, is critical for high-quality text-to-image~\cite{rombach2022high} and text-to-3D~\cite{Poole2022DreamFusionTU} generation.

Despite its popularity, we think a solid understanding of  \cfgguide~is missing.
Recently, several efforts provide insights by studying \cfgguide~from a theoretical perspective~\cite{Bradley2024ClassifierFreeGI, Xia2024RectifiedDG,Chidambaram2024WhatDG} showing that sampling from \cfgguide~is not the same as sampling from a sharpened distribution.

Instead of solely focusing on \cfgguide~as done in the works mentioned above, we trace back to the root of  \cfgguide,~\ie,~\cgguide~\cite{dhariwal2021diffusion}.
It is \cgguide~that decomposes the \textit{conditional} generation into a combination of an \textit{unconditional} generation and a classifier prediction.
\Cfgguide~adopts this decomposition and replaces the classifier by randomly dropping conditioning during training~\cite{ho2022classifier}.
This connection motivates us to carefully study \cgguide's derivation and its behavior.

We first identify a key assumption underlying \cgguide's decomposition, which is also central to \cfgguide~due to the connection mentioned above, that often fails to hold.
This issue results in different behaviors for 1) a vanilla denoising diffusion conditional generation; and 2) a generation that follows the decomposition of \cgguide~as well as \cfgguide.
On synthetic 1D data, the vanilla conditional generative model produces straight denoising paths while the decomposed version results in distorted trajectories that are \textit{pushed away from the classifier's decision boundary}.
This discrepancy is exacerbated  with the commonly used large guidance scale.

The above observation motivates us to further study the sensitivity of \cgguide~to the accuracy of the \textit{classifier}.
We find that \cgguide~generations are dominated by the behavior of the classifier that provides guidance.
In other words, conditional generation via \cgguide~is achieved via pushing the generation away from the class decision boundaries.
A similar observation is obtained for \cfgguide~as well.
To further verify this classifier-centric perspective, we study a postprocessing step to push samples from the trained model, \textit{mainly around the decision boundaries}, to their nearest neighbors in the real data.
Experiments on various datasets demonstrate the improvement of generation, verifying our understanding.

In summary, our contribution is a systematic empirical study of both classifier and classifier-free guidance from a classifier-centric perspective for intuitive understandings.

\section{Related Works}

\noindent\textbf{Trading diversity for fidelity in conditional generation} is a long-standing problem that has been actively studied by the community.
For probabilistic models trained with the maximum likelihood objective, \citet{Ackley1985ALA} propose low-temperature sampling to effectively focus on the mode of the learned distribution, borrowing ideas from statistical mechanics~\cite{metropolis1953equation}.
This technique has also been employed beneficially for high-quality image synthesis~\cite{Parmar2018ImageT,Kingma2018GlowGF}.
Recent large language models (LLMs)~\cite{achiam2023gpt,dubey2024llama} also exploit this idea, balancing creativity and determinism via temperature control during next token prediction via the learned probability model~\cite{Brown2020LanguageMA}.
For image synthesis with generative adversarial nets (GANs), the truncation trick~\cite{Brock2018LargeSG,Karras2018ASG} was developed to enforce sampling from a truncated normal distribution rather than the standard normal prior.
This encourages conditional generations to remain close to the mode of the data distribution observed during training, preventing them from diverging too far.
More recently, denoising diffusion models have demonstrated impressive generation capabilities in various domains~\cite{Kong2020DiffWaveAV,Poole2022DreamFusionTU,rombach2022high,Chen2020WaveGradEG}.
\Cfgguide~\cite{ho2022classifier}, built upon \cgguide~\cite{dhariwal2021diffusion}, has emerged as a standard for controlling conditional generations of denoising diffusion models.
Our work contributes to the understanding of the trade-off between diversity and fidelity in the field of denoising diffusion models via carefully studying classifier and classifier-free guidance from a classifier-centric perspective.

\noindent\textbf{Generation with guidance} is closely related to our study.
Techniques discussed in the preceding paragraph, except \cgguide, solely require trained generative models,~\eg,~the generator in GANs, to control the diversity and fidelity trade-off.
In contrast, guidance relies on a separate model to influence the conditional generation.
Rejection sampling~\cite{casella2004generalized} is an active area of research in this direction.
For GANs, prior works use the discriminator paired with the generator to reject generations for which the discriminator has high confidence~\cite{Azadi2018DiscriminatorRS,Turner2018MetropolisHastingsGA}.
Alternatively,~\citet{Che2020YourGI} utilize the discriminator to reject samples in the latent space.
For variational autoencoders (VAEs)~\cite{Kingma2013AutoEncodingVB}, learnable acceptance functions have been studied for both  prior~\cite{Bauer2018ResampledPF,AnejaNEURIPS2021} and posterior~\cite{Grover2018VariationalRS,Jankowiak2023ReparameterizedVR} rejection sampling.
Other works explore pre-trained classifiers to provide guidance.
~\citet{Razavi2019GeneratingDH} use a classifier trained on ImageNet~\cite{deng2009imagenet} to reject samples that cannot be well-recognized.
Thanks to the recent progress of representation learning, several prior works exploit CLIP~\cite{Radford2021LearningTV} to provide guidance on conditional generations with GANs~\cite{Galatolo2021GeneratingIF,Patashnik2021StyleCLIPTM}.
\citet{Kim2022RefiningGP} improves the quality of a pre-trained model via refining denoising trajectories with guidance from a discriminator that distinguishes between real and fake denoising paths.
~\citet{Dinh2023PixelAsParamAG, Dinh2023RethinkingCD} mitigates the conflicts between quality and diversity caused by the guidance from a gradient and progressive perspective.
\Cgguide~\cite{dhariwal2021diffusion} steers generation using a classifier trained along the denoising path. Inspired by this, we take a classifier-centric view to intuitively understand the behaviors of \cgguide~and \cfgguide.

\noindent\textbf{\Cfgguide} has attracted more and more attention in the community.
Theoretically, several recent works clarify that sampling with \cfgguide~does not correspond to sampling from a tilted distribution~\cite{Bradley2024ClassifierFreeGI, Xia2024RectifiedDG,Chidambaram2024WhatDG,Wu2024TheoreticalIF}, a misconception that is popular in the community.
~\citet{Bradley2024ClassifierFreeGI} further prove that \cfgabb~is equivalent to the predictor-corrector mechanism~\cite{Song2020ScoreBasedGM} in the continuous-time limit.
Empirically, prior works improve generation quality by refining \cfgguide.
\citet{Sadat2023CADSUT} dynamically adjust the scale of \cfgguide~to improve the generation diversity.
\citet{Lin2023DiffusionMW} argue that \cfgguide~essentially behaves as a perceptual loss and propose to incorporate a self-perceptual objective during training.
\citet{chung2024cfgplusplus} introduce CFG++ to mitigate the issue of an off-manifold denoising path via a refined sampling formulation and a small scale.
Different from these works that solely focus on \cfgguide, we instead trace back to the origin,~\ie,~\cgguide~\cite{dhariwal2021diffusion}.
We systematically study the role the classifier plays in the performance of \cgguide~and found \cgguide~essentially pushes the generation away from the decision boundary. %
Furthermore, we demonstrate that this is also true for~\cfgguide.

\section{Analysis}\label{sec: analysis}

\subsection{Conditional Generation as Denoising Diffusion}

\begin{figure}[!t]
    \centering
    \includegraphics[width=0.8\columnwidth]{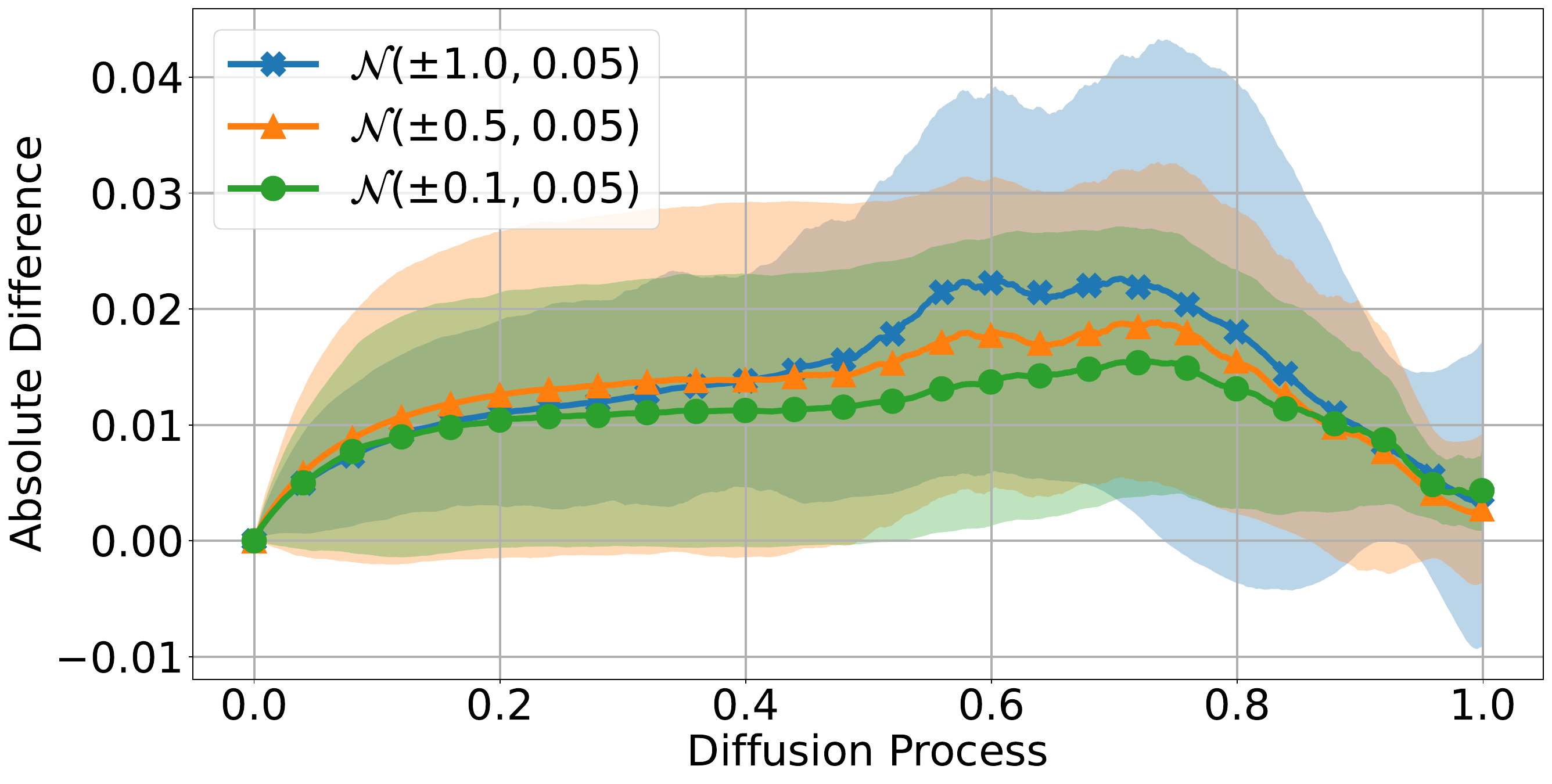}
    \caption{
      \textbf{\Cgguide~decomposition (\equref{eq: guided_diffusion}) does not always hold.} 
      We apply \cgguide~on 1D data from $\mathcal{N}(\pm 1.0, 0.05)$, $\mathcal{N}(\pm 0.5, 0.05)$, and $\mathcal{N}(\pm 0.1, 0.05)$ respectively.
      The denoising diffusion process starts from left to right.
      For each dataset, we train a vanilla conditional diffusion model and a decomposed version,~\ie,~an unconditional diffusion model and a classifier. %
      We generate 20k samples (10k for each class) from both sides of~\equref{eq: guided_diffusion} with the same initial noises and compute the absolute differences for each step in the denoising diffusion process.
      This plot shows the average as well as the standard deviation for the difference.
      Apparently, the \cgguide~decomposition doesn't hold with equality.
    }
    \label{fig: gaussian traj diff cg}
\end{figure}

The goal of conditional generation is to sample the data of interest $\data_0$, \eg, images, from a conditional distribution,~\ie,~$\data_0 \sim p_\theta(\data_0 \vert \cond)$. Here, $\cond$ is the conditioning information, \eg, class labels.
Note, hereafter we use $\theta$  to subsume all learnable parameters for simplicity.

In this work, we focus on denoising diffusion models~\cite{song2019gen,vincent2011connection,Ho2020DenoisingDP}.
A denoising diffusion process generates data from white noise by introducing a sequence of latent variables $\data_{1:T} \triangleq \{\data_1, \dots, \data_T\}$ that form a Markov chain $p_\theta (\data_0 \vert \cond) = \int p_\theta (\data_0, \data_{1:T} \vert \cond) \,d \data_{1:T} \triangleq \int p(\data_T \vert c) \prod_{t=0}^{T-1} p_\theta (\data_{t} \vert \data_{t+1}, \cond) \,d \data_{1:T}$
\begin{align}
    \approx &\int p(\data_T) \prod_{t=0}^{T-1} p_\theta (\data_{t} \vert \data_{t+1}, \cond) \,d \data_{1:T}. \label{eq: reverse ddpm}
\end{align}
The last step is due to  $p(\data_T \vert \cond)$ being almost identical to an isotropic Gaussian, independent of the condition $\cond$.

Following DDPM~\cite{Ho2020DenoisingDP}, $p_\theta (\data_{t} \vert \data_{t+1}, \cond)$ is defined as a Gaussian $\mathcal{N}(\data_t; \mu_\theta(\data_{t+1}, t+1, \cond), (1 - \alpha_{t+1}) \mathbf{I})$, trying to reverse a forward diffusion process.
Here $\{\alpha_t\}_{t=1}^T$ is a pre-defined schedule for the forward diffusion process and $\mu_\theta(\data_{t+1}, t+1, \cond)$ is tasked to predict the corresponding $\data_t$ in the forward diffusion process.  
Specifically, the forward diffusion process gradually corrupts the clean data $\data_0$ with Gaussian noise: $\data_{t+1} \sim q(\data_{t+1} \vert \data_t) \triangleq \mathcal{N} (\data_{t+1}; \sqrt{\alpha_{t+1}} \,\data_t, (1 - \alpha_{t+1}) \mathbf{I}), \forall t \in \{0, \dots, T-1\}$.
Notably, with $\bar{\alpha}_t \triangleq \prod_{s=1}^t \alpha_s$, we have $q(\data_t \vert \data_0) = \mathcal{N} ( \data_t; \sqrt{\bar{\alpha}_t} \,\data_0, (1 - \bar{\alpha}_t) \mathbf{I}) = \sqrt{\bar{\alpha}_t} \,\data_0 + \sqrt{1 - \bar{\alpha}_t} \cdot \noise$, where $\noise \sim \mathcal{N}(\mathbf{0}, \mathbf{I})$.
Therefore,~\citet{Ho2020DenoisingDP} propose to reduce the learning of $\mu_\theta(\data_{t+1}, t+1, \cond)$ to predicting the noise with $\noise_\theta(\data_{t+1}, t+1, \cond)$ as we have $\mu_\theta(\data_{t+1}, t+1, \cond) =$
\begin{align}
     \frac{1}{\sqrt{\alpha_t}} \left( \data_{t+1} - \frac{1 - \alpha_t}{\sqrt{1 - \bar{\alpha}_t}} \, \noise_\theta(\data_{t+1}, t+1, \cond) \right).
\end{align}

Leveraging the link between denoising diffusion and score matching~\cite{Song2020ScoreBasedGM,vincent2011connection}, we have
\begin{align}
    \noise_\theta(\data_{t}, t, \cond) = - \sqrt{1 - \bar{\alpha}_t} \; \nabla_{\data_{t}} \log p_\theta (\data_{t} \vert \cond). \label{eq: score matching}
\end{align}

\subsection{Classifier Guidance Revisited}\label{sec: cg}

\citet{dhariwal2021diffusion} propose \textit{\cgguide} to decompose the conditional denoising diffusion process in~\equref{eq: reverse ddpm} as follows:
\begin{align}
    p_\theta (\data_{t} \vert \data_{t+1}, \cond) = Z \, p_\theta (\data_{t} \vert \data_{t+1}) \, p_\theta ( \cond \vert \data_{t}). \label{eq: guided_diffusion}
\end{align}
$Z$ is a normalizing factor independent of $\data_{t}$.
$p_\theta (\data_{t} \vert \data_{t+1})$ is an unconditional denoising diffusion process and 
$p_\theta( \cond \vert \data_{t})$ is a classifier used to predict the probability that $\data_t$ aligns with the conditioning information $\cond$.
Note,\textit{~\equref{eq: guided_diffusion} is not a trivial Bayes expansion}. See~\secref{supp sec: proof} for more.

Revisiting the derivation of~\equref{eq: guided_diffusion}, the key step reduces to the following definition (see~\equref{supp eq: key assumption}):
\begin{align}
    \hat{q}(\data_{t+1} \vert \data_t, \cond) \triangleq q(\data_{t+1} \vert \data_t). \label{eq: guided_diffusion_key_step}
\end{align}
Note, $\hat{q}(\data_{t+1} \vert \data_t, \cond)$ is a newly-defined conditional forward diffusion process.
At a high level,~\equref{eq: guided_diffusion_key_step} tries to convert any \textit{conditional} forward diffusion process,~\ie,~$\hat{q}(\data_{t+1} \vert \data_t, \cond)$, into an \textit{unconditional} forward diffusion process,~\ie,~$q(\data_{t+1} \vert \data_t)$.
Based on~\equref{eq: guided_diffusion_key_step},~\citet{dhariwal2021diffusion} derive that the reverse process of $\hat{q}(\data_{t+1} \vert \data_t, \cond)$ can be decomposed into a combination of an unconditional denoising diffusion process and a classifier prediction as in~\equref{eq: guided_diffusion}.

\begin{figure*}[!t]
    \vspace{0pt} 
    \centering
    \begin{minipage}[t]{0.49\linewidth}
        \begin{subfigure}[t]{\linewidth}
            \centering
            \adjincludegraphics[width=\linewidth,trim={{0.0\width} 0 {0\width} 0},clip]{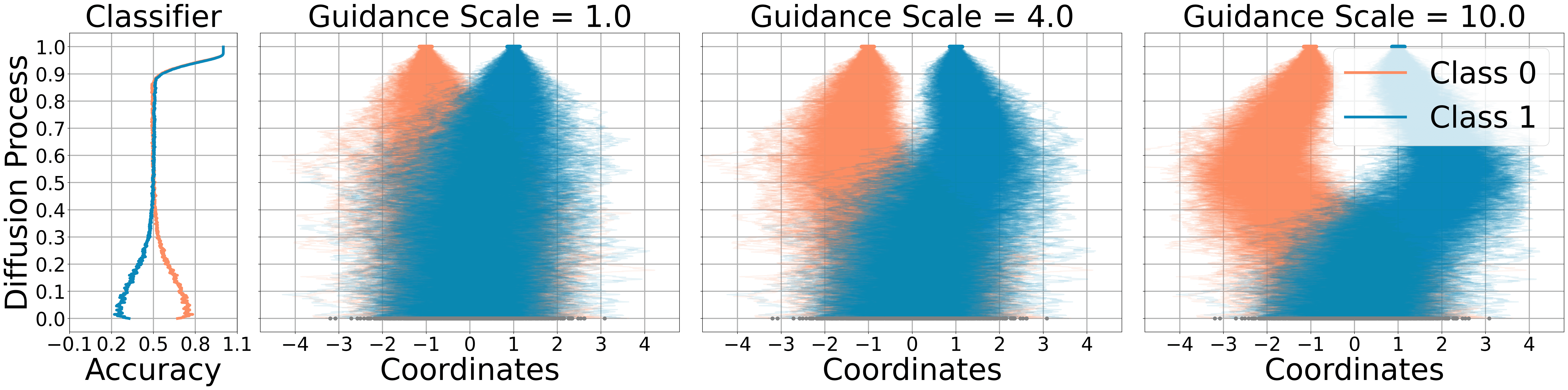}
            \captionsetup{width=\linewidth}
            \caption{
              \textbf{\Cgguide~with a nonlinear classifier.}
            }
            \label{fig: gaussian traj cg classifier nonlinear}
        \end{subfigure}
    \end{minipage}%
    \hfill
    \begin{minipage}[t]{0.49\linewidth}
        \begin{subfigure}[t]{\linewidth}
            \centering
            \adjincludegraphics[width=\linewidth,trim={{0.0\width} 0 {0\width} 0},clip]{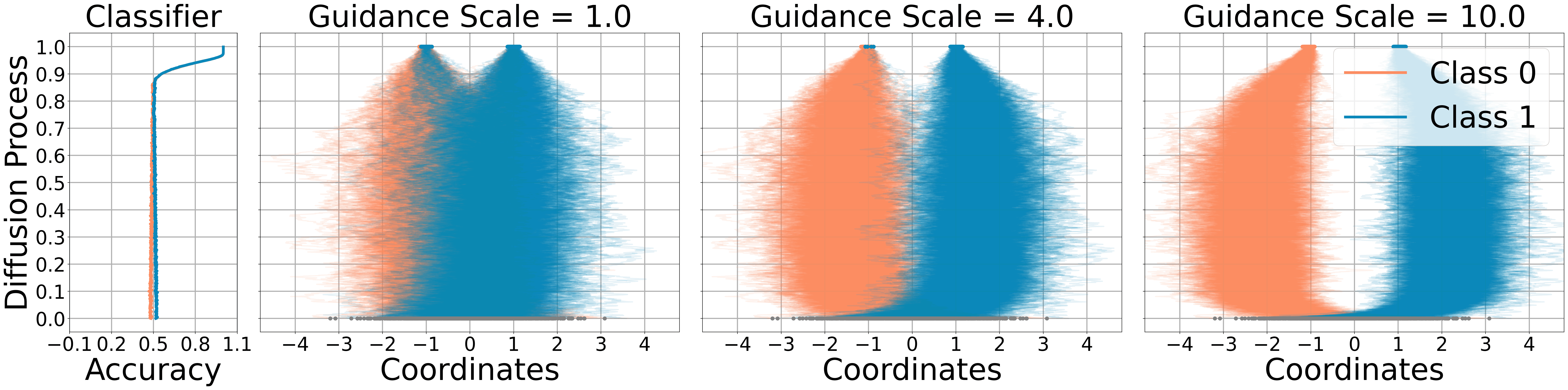}
            \captionsetup{width=\linewidth}
            \caption{
              \textbf{\Cgguide~with a linear classifier.}
            }
            \label{fig: gaussian traj cg classifier linear}
        \end{subfigure}
    \end{minipage}%
    \caption{
      \textbf{\Cgguide~behavior is dominated by the classifier.}
      We apply denoising diffusion models with \cgguide~on a 1D dataset with data from $\mathcal{N}(\pm 1.0, 0.05)$.
      The classifiers in~\figref{fig: gaussian traj cg classifier nonlinear} and~\ref{fig: gaussian traj cg classifier linear}  differ.
      The denoising diffusion process for all plots starts from the bottom to the top.
      In~\figref{fig: gaussian traj cg classifier nonlinear} and~\ref{fig: gaussian traj cg classifier linear}, the first plot demonstrates the classifier's accuracy on a validation set for each class through the diffusion process,~\ie,~$p_\theta ( \cond \vert \data_{t})$ in~\equref{eq: guided_diffusion}, while the remaining three plots display the diffusion trajectories with different guidance scales.
      We observe:
      1) \cgguide~essentially pushes the diffusion process away from the classifier's decision boundary that is around the origin;
      and 2) different classifiers can produce entirely different trajectories (\figref{fig: gaussian traj cg classifier nonlinear}~\vs~\ref{fig: gaussian traj cg classifier linear}).
      Since we use the same initial noise and the same unconditional diffusion model,~\ie,~$p_\theta (\data_{t} \vert \data_{t+1})$ in~\equref{eq: guided_diffusion}, for all plots,  differences are solely due to the classifier.
    }
    \label{fig: gaussian traj cg}
\end{figure*}

However, it is questionable whether the assumption of~\equref{eq: guided_diffusion_key_step} always holds: why should a conditional denoising process behave identical to an unconditional one? If \equref{eq: guided_diffusion_key_step} does not hold everywhere, the two sides in~\equref{eq: guided_diffusion} may differ too.
Indeed, our experiments on synthetic 1D data verify our suspicion as shown in~\figref{fig: gaussian traj diff cg}.
Furthermore, not only does the vanilla conditional model (left side of~\equref{eq: guided_diffusion})  behave differently from the proposed decomposition (right side of~\equref{eq: guided_diffusion}), but different instantiations of the classifier $p_\theta ( \cond \vert \data_{t})$ will produce significantly divergent behaviors as well.
The ``guidance scale = 1'' plots in~\figref{fig: gaussian traj cg classifier nonlinear} and~\ref{fig: gaussian traj cg classifier linear} clearly illustrate this.

\noindent\textbf{Large \cgguide~scale $w$} introduced by~\citet{dhariwal2021diffusion} will amplify the difference demonstrated above.
Specifically,~\citet{dhariwal2021diffusion} suggest increasing the impact of the classifier with $w > 1$ and sampling from a distribution that is skewed towards high classifier confidence:
\begin{align}
    \data_t \sim p_\theta (\data_{t} \vert \data_{t+1}) \, p_\theta ( \cond \vert \data_{t})^w. \label{eq: cg scale}
\end{align}
Similar to~\equref{eq: score matching},~\citet{dhariwal2021diffusion} show that~\equref{eq: cg scale} can be re-formulated such that $\data_t$ can be sampled via predicting the following noise $\tilde{\noise}_\theta (\data_t, t, \cond) \triangleq$
\begin{align}
     \noise_\theta (\data_t, t) - w \cdot \sqrt{1 - \bar{\alpha}_t} \;\nabla_{\data_t} \log p_\theta (\cond \vert \data_t), \label{eq: cg noise format}
\end{align}
where $\noise_\theta (\data_t, t)$ is the corresponding noise estimator for the unconditional denoising diffusion process $p_\theta (\data_{t} \vert \data_{t+1})$.
As shown in~\figref{fig: gaussian traj cg}, \cgguide's behaviors are dominated by the characteristics of the classifier.

It is worth noticing  that \cgguide~with larger and larger guidance scales continuously distorts the straight-like denoising diffusion trajectories to \textit{push them away from the classifier's decision boundary}.
In other words, the goal of \textit{conditional} generation via \cgguide~is achieved by avoiding those areas in which the classifier is uncertain.
This  explains why a large guidance scale can produce high-fidelity images compared to results generated with a low guidance scale.
The reason is that a low guidance scale is not strong enough to move the diffusion trajectories away from areas on the data distribution manifold where different conditional information intersect.
Due to the entanglement, these areas naturally form the decision boundary for a \textit{well-trained} classifier.
With a large guidance scale and a well-trained classifier, \cgguide~can completely avoid ambiguous areas on the image manifold and generate results \textit{unambiguously} aligned with the conditional information.

The obvious next question:
can this reasoning for \cgguide~be generalized to \cfgguide?

\begin{figure}[!t]
    \centering
    \includegraphics[width=\linewidth]{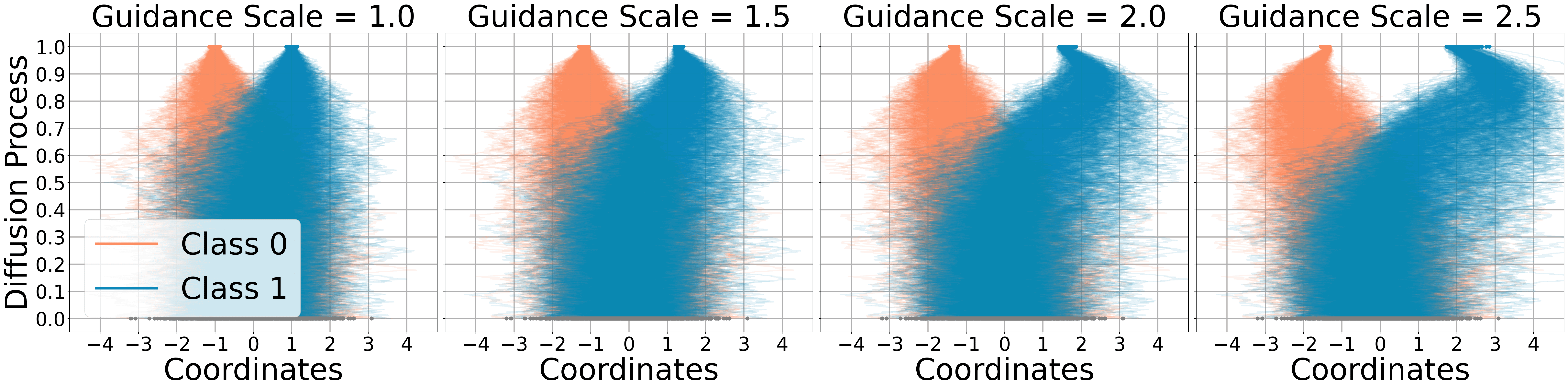}
    \caption{
      \textbf{\Cfgguide~distorts denoising diffusion trajectories.} 
      We apply denoising diffusion models with \cfgguide~on a 1D dataset composed of data from $\mathcal{N}(\pm 1.0, 0.05)$.
      The denoising diffusion process for all plots starts from the bottom to the top.
      We use the same trained model as well as the same initial noise for all plots.
      The trajectory differences are solely caused by different guidance scales.
      Different scales in~\figref{fig: gaussian traj cg} and this figure arise from \cgguide~and \cfgguide's differing sensitivities. Here, scale=2.5 distorts trajectories significantly, while~\figref{fig: gaussian traj cg}’s scale=4 causes minor changes. We hypothesize that \cfgguide's greater sensitivity stems from its training with conditioning dropout.
    }
    \label{fig: gaussian traj cfg}
\end{figure}

\subsection{Classifier-Free Guidance Revisited}\label{sec: cfg}

\Cfgguide~was introduced to eliminate the reliance on a separate classifier~\cite{ho2022classifier}.
Intuitively, with Bayes rule, we have $p(\cond \vert \data_t) = p(\cond) \, p(\data_t \vert \cond) / p(\data_t)$.
Consqeuently, $\nabla_{\data_t} \log p_\theta (\cond \vert \data_t)$ can be decomposed as $\nabla_{\data_t} \log p_\theta (\data_t \vert \cond) - \nabla_{\data_t} \log p_\theta (\data_t)$, where $p(c)$ disappears as it is independent of $\data_t$.
When substituting this into~\equref{eq: cg noise format} and considering~\equref{eq: score matching}, we have $\tilde{\noise}_\theta (\data_t, t, \cond) =$
\begin{align}
    &\noise_\theta (\data_t, t) + w \cdot \left( \noise_\theta(\data_{t}, t, \cond) - \noise_\theta(\data_{t}, t)  \right). \label{eq: cg noise format bayes}
\end{align}
The effect of \cgguide~can be achieved by training two noise estimators for both conditional ($\noise_\theta(\data_{t}, t, \cond)$) and unconditional ($\noise_\theta(\data_{t}, t)$) denoising diffusion processes respectively.
In practice,~\citet{ho2022classifier} propose to only train one conditional denoising diffusion model but randomly drop out the conditioning information $\cond$ during training to mimic the unconditional process.

\begin{figure*}[!ht]
    \vspace{0pt} 
    \centering
    \begin{minipage}[t]{0.865\textwidth}
        \begin{subfigure}[t]{\textwidth}
            \centering
            \adjincludegraphics[width=\textwidth,trim={{0.0\width} 0 {0\width} 0},clip]{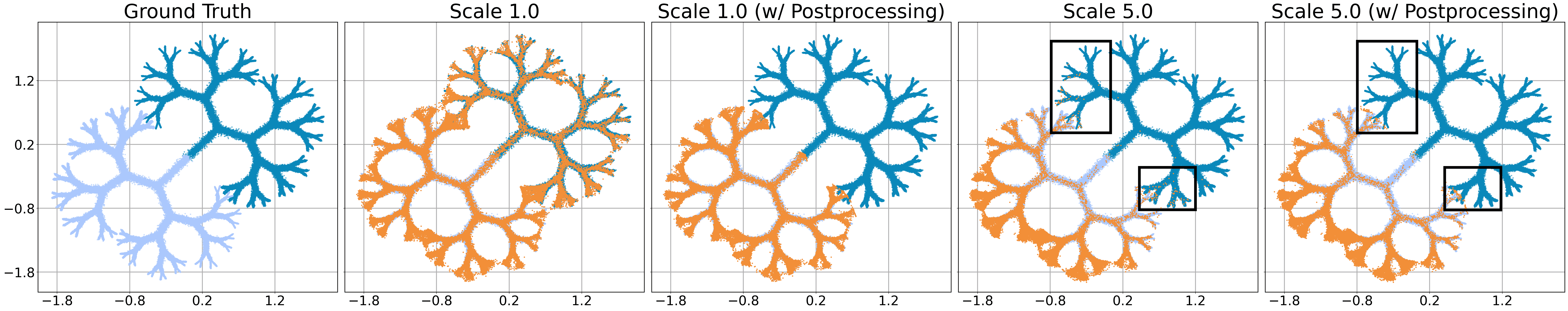}
            \captionsetup{width=\textwidth}
            \caption{
              \textbf{Low entanglement.}
            }
            \label{fig: fractal cg 0.0}
        \end{subfigure}
    \end{minipage}%
    \hfill
    \begin{minipage}[t]{0.865\textwidth}
        \begin{subfigure}[t]{\textwidth}
            \centering
            \adjincludegraphics[width=\textwidth,trim={{0.0\width} 0 {0\width} 0},clip]{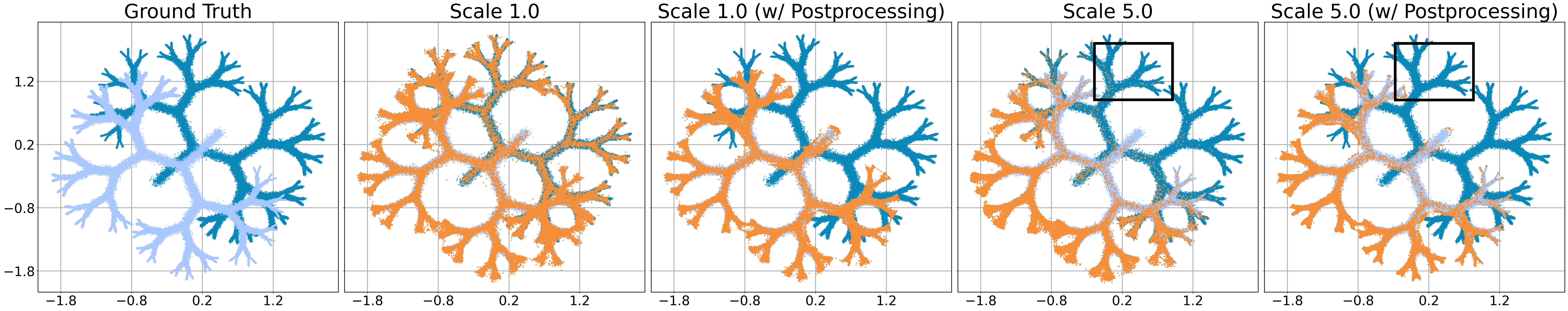}
            \captionsetup{width=\textwidth}
            \caption{
              \textbf{Medium entanglement.}
            }
            \label{fig: fractal cg 0.5}
        \end{subfigure}
    \end{minipage}%
    \hfill
    \begin{minipage}[t]{0.865\textwidth}
        \begin{subfigure}[t]{\textwidth}
            \centering
            \adjincludegraphics[width=\textwidth,trim={{0.0\width} 0 {0\width} 0},clip]{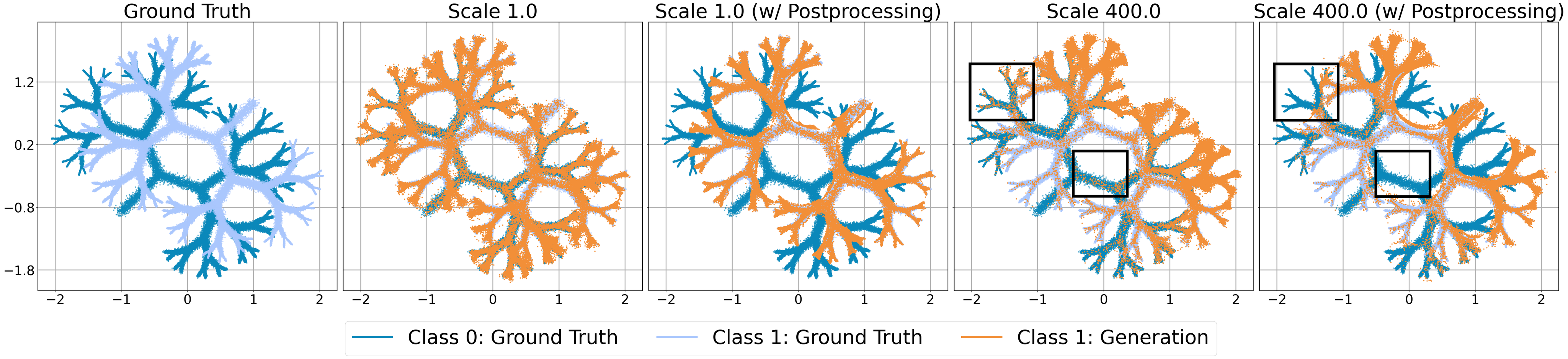}
            \captionsetup{width=\textwidth}
            \caption{
              \textbf{High entanglement.}
            }
            \label{fig: fractal cg 1.0}
        \end{subfigure}
    \end{minipage}%
    \caption{
      \textbf{\Cgguide~with flow-matching based postprocessing (\secref{sec: postprocess}) on 2D fractal data.}
      After training, all three classifiers' decision boundaries roughly align with the diagonal from top-left to bottom-right.
      See~\secref{sec: fractal exp setup} for the experiment setup.
      In~\figref{fig: fractal cg 0.0} to~\ref{fig: fractal cg 1.0}, the $3^\text{rd}$ (and $5^\text{th}$) plot show generated samples after applying postprocessing  on generations from the $2^\text{nd}$ (and $4^\text{th}$) plot.
      For a clear visualization, we only display generations for one class (see~\figref{fig: fractal cg class 0} for the other class).
    }
    \label{fig: fractal cg}
\end{figure*}

One may notice that~\equref{eq: guided_diffusion}
\textit{lays the foundation for  \cfgguide~\cite{ho2022classifier}}.
If this is not clear, please refer to~\secref{supp sec: proof} for more details.
We want to know whether the connection between~\cgguide~and~\cfgguide~can be used to show that \cfgguide~inherits the characteristics of \cgguide.
Namely, does \cfgguide~also try to push the diffusion trajectories away from the data's decision boundary?
Note, \cfgguide~does not directly involve any explicit classifier.
However, based on our discussion in~\secref{sec: cg}, a well-trained classifier's decision boundary naturally aligns with the data's decision boundary.
Experiments on 1D synthetic datasets provide an affirmative answer as shown in~\figref{fig: gaussian traj cfg}.

\subsection{Verification on High-Dimensional Data}
\label{sec: postprocess}

So far, we have developed and verified our understanding on synthetic 1D data thanks to clear visualizations.
A natural question arises: how can we assess our understanding on high-dimensional data, where visualizing the decision boundary, let alone the denoising trajectory, is non-trivial? To address this, we propose an alternative approach that perturbs samples away from the decision boundary. This allows us to \textit{indirectly} evaluate our classifier-centric understanding.

If our classifier-centric understanding is correct,~\ie, if~\cgguide~and~\cfgguide~perform conditional generation by moving samples away from the decision boundary, then low-quality generations should occur more frequently near the decision boundary, due to its complex structure in high-dimensional spaces.
For a pre-trained generative model, if we can construct a postprocessing step that moves samples away from the decision boundary, the overall quality of the generation should improve.

For this, we develop a generic postprocessing step to help verify our understanding on high-dimensional data.
Let $\setreal$ refer to a set of samples from the real data distribution while $\set$ denotes a set of generated samples from a generative model.
Note, $\set$ is agnostic to the specific choice of generation strategy,~\eg,~it does not matter whether $\set$ was produced with \cgguide~or \cfgguide.
Our goal is to move the distribution underlying $\set$ closer to the distribution represented by $\setreal$ mainly around the decision boundary.
We train a rectified flow~\cite{Liu2022FlowSA,Lipman2022FlowMF} $v_\theta$ via
\begin{align}
    &\min_{v_\theta} \! \int_0^1 \! \mathbb{E}_{\set} \! \left[ \Vert (\hat{\data} \!-\! \texttt{NN}(\hat{\data}, \setreal)) - v_\theta(\hat{\data}_t, \cond, t) \Vert^2 \right] dt, \label{eq: post flow}
\end{align}
where $\hat{\data} \sim \set, \,\hat{\data}_t = (1 - t) \cdot \hat{\data} + t \cdot \texttt{NN}(\hat{\data}, \setreal)$.
Here $\texttt{NN}(\hat{\data}, \setreal)$ represent the nearest neighbor sample for $\hat{\data}$ in the real data set $\setreal$.

\begin{figure*}[!t]
    \centering
    \includegraphics[width=\linewidth]{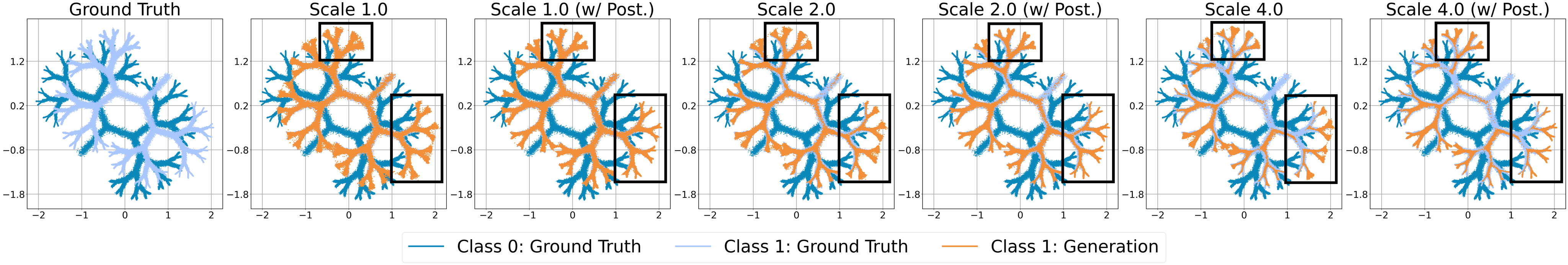}
    \caption{
      \textbf{\Cfgguide~with flow-matching based postprocessing (\secref{sec: postprocess}) on 2D fractal data.}
      The level of entanglement is the same as that in~\figref{fig: fractal cg 1.0}.
      Each plot with `Post.` in the title displays generations after applying postprocessing on samples from the corresponding previous plot.
      We observe that the postprocessing step continuously improves the fidelity of the generations via moving samples around the decision boundary back to the real data distribution as the leaf branches become much sharper, regardless of the guidance scale we use.
    }
    \label{fig: fractal cfg}
\end{figure*}

\begin{table}[!t]
\setlength\aboverulesep{0pt}

\captionsetup{width=\linewidth}
\caption{\textbf{Nearest neighbor distance between generations and ground truth for~\figref{fig: fractal cfg}.}
We report average nearest neighbor (\texttt{NN}) distance ($\times 10^{-5}$) for 20k generations from the same noise, formatted as $A / B / C$: before postprocessing (A) / postprocessed with nearest (B) / postprocessed with random sampling from 20 candidates (C).
B~\vs~C shows random sampling outperforms picking the nearest.
}
\label{tab: fractal nearest dist}
\begin{adjustbox}{width=\linewidth,center}
\setlength{\tabcolsep}{0.2cm}
{
\begin{tabular}{lrrr}
\toprule
Guidance Scale  & $w=1$ &  $w=2$ &  $w=4$ \\
\hline
Class 0  & 5.95 / 3.11 / 1.57 & 5.31 / 1.70 / 1.01 & 11.7 / 2.05 / 1.12 \\
Class 1  & 12.4 / 3.25 / 1.70 & 6.09 / 1.88 / 1.00 & 6.51 / 1.41 / 0.79 \\
\bottomrule
\end{tabular}
}
\end{adjustbox}
\end{table}

We emphasize the use of $\texttt{NN}$ in \equref{eq: post flow}, which differs from classic rectified flow formulations.
Importantly, the use of $\texttt{NN}$ automatically balances between 1) already-high-quality generations; and 2) low-quality generations.
Based on our study in~\secref{sec: cg} and~\secref{sec: cfg}, the training will focus on generations around decision boundaries where low-quality generations usually occur.
Concretely, if $\hat{\data}$ is already a high-fidelity generation,~\ie,~close to $\setreal$, $\hat{\data} - \texttt{NN}(\hat{\data}, \setreal)$ will be extremely small, providing a negligible learning signal.
In practice, inspired by~\cite{Tong2023ImprovingAG}, we do not always use the nearest neighbor $\texttt{NN}(\hat{\data}, \setreal)$.
Instead, we first find top-$k$ nearest neighbors and randomly select one from the top-$k$ as the target during each training iteration. This randomness provides more opportunities to avoid local optima.

After training the postprocessing flow,  conditional generation  involves two steps:
1) sampling from the original denoising diffusion model $p_\theta(\data_0 \vert \cond)$ in~\equref{eq: reverse ddpm} to obtain a sample $\hat{\data}_0$;
and 2) running an ODE solver over the time interval $[0, 1]$ to solve $d \sample_t/dt = v_\theta (\sample_t, \cond, t)$ numerically, starting from $\sample_0 = \hat{\data}_0$.
The ODE solver's output $\sample_1$ is the final generation.
We use $\hat{\data}_0$ for the base model output and $\sample_t$ to emphasize that postprocessing is based on a separate flow matching procedure.

The proposed postprocessing is related to autoguidance~\cite{Karras2024GuidingAD}, which guides the model training with a bad version of itself.
Autoguidance moves samples in the direction given by the difference between an inferior version and the current model. 
In contrast, our postprocessing flow model is based on a pre-trained model and real data.
More importantly, we propose the postprocessing step primarily to verify our classifier-centric understanding.

\begin{figure*}[!t]
    \vspace{0pt} 
    \centering
    \begin{minipage}[t]{0.4\textwidth}
        \begin{subfigure}[t]{0.49\textwidth}
            \centering
            \adjincludegraphics[width=\textwidth,trim={{0.0\width} 0 {0.2\width} 0},clip]{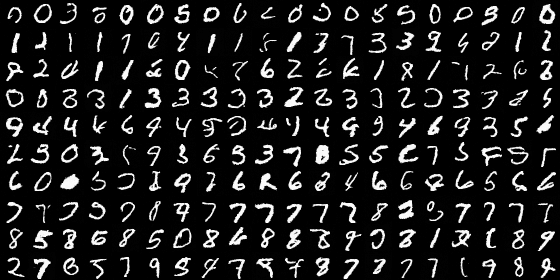}
            \captionsetup{width=\textwidth}
            \caption{
              \textbf{Generations before postprocessing step.}
            }
            \label{fig: mnist cg before}
        \end{subfigure}
        \hfill
        \begin{subfigure}[t]{0.49\textwidth}
            \centering
            \adjincludegraphics[width=\textwidth,trim={{0.0\width} 0 {0.2\width} 0},clip]{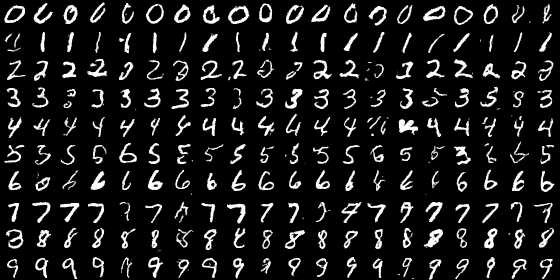}
            \captionsetup{width=\textwidth}
            \caption{
              \textbf{Generations after postprocessing step.}
            }
            \label{fig: mnist cg after}
        \end{subfigure}
    \end{minipage}%
    \hfill
    \begin{minipage}[t]{0.59\textwidth}
        \begin{subfigure}[t]{\textwidth}
            \centering
            \adjincludegraphics[width=\textwidth,trim={{0.0\width} 0 {0\width} 0},clip]{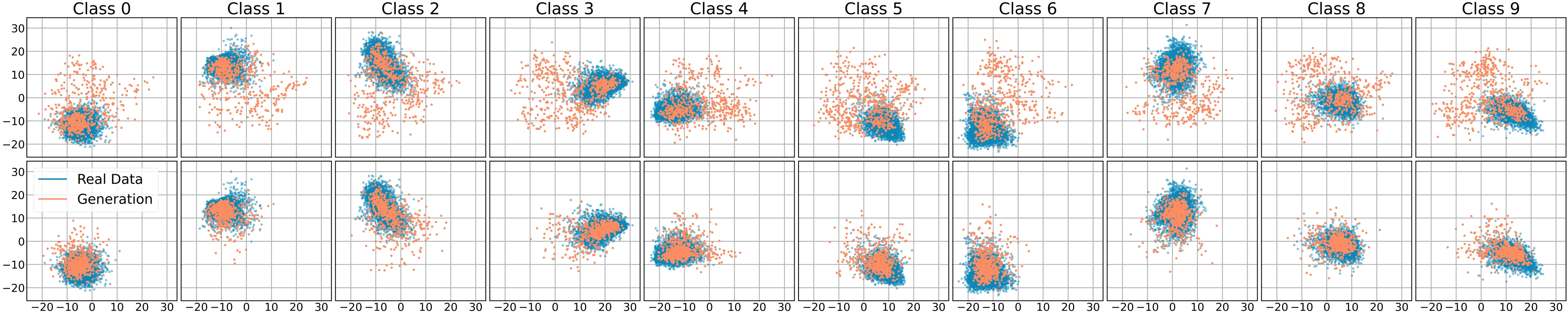}
            \captionsetup{width=\textwidth}
            \caption{
              \textbf{Postprocessing moves generations around the decision boundaries closer to the real data distribution.}
            }
            \label{fig: mnist cg pca}
        \end{subfigure}
    \end{minipage}%
    \caption{
      \textbf{\Cgguide~(scale 1.0) with flow-matching based postprocessing (\secref{sec: postprocess}) on MNIST.}
      \figref{fig: mnist cg before} and~\ref{fig: mnist cg after} share the same initial noises for corresponding cells.
      Conditioning information from top to bottom row is the digit 0 to 9.
      ~\figref{fig: mnist cg pca} shows that the flow matching based postprocessing clearly improves the alignment between generations and conditioning:
      The top and bottom rows in~\figref{fig: mnist cg pca} correspond to~\figref{fig: mnist cg before} and~\ref{fig: mnist cg after} respectively.
      The postprocessing moves the learned distribution (orange clusters) closer to the real one (blue clusters).
      See~\secref{supp sec: mnist exp setup} for the experiment details.
    }
    \label{fig: mnist cg}
\end{figure*}

\section{Experiments}\label{sec: exp}

\subsection{2D Fractal}\label{sec: 2d fractal}

The 2D Fractal dataset is represented by a mixture of Gaussians, similar to the dataset used by~\citet{Karras2024GuidingAD}. Please refer to~\secref{sec: fractal construct} for details.
As displayed in ``Ground Truth'' plots in~\figref{fig: fractal cg}, \textit{this synthetic dataset provides an easy way to control the level of entanglement among data from different classes}.
We use this dataset to verify
1) our analysis in~\secref{sec: analysis} on 1D data can be generalized;
and 2) the postprocessing step in~\secref{sec: postprocess} is an effective proxy to assess our classifier-centric understanding.

We qualitatively illustrate the results for \cgguide~and \cfgguide~in~\figref{fig: fractal cg} and~\figref{fig: fractal cfg} respectively.
We choose top-$20$ nearest neighbors when training the rectified flow in the experiments.
As can be seen clearly, the larger the guidance scale, the further are the samples from the decision boundaries, corroborating our classifier-centric understanding.
Further, the proposed postprocessing step mainly moves samples around the decision boundaries while keeping generations that are already close to real data untouched.
Quantitatively,~\tabref{tab: fractal nearest dist} verifies the effectiveness of our approach for~\figref{fig: fractal cfg}.

Additionally, according to~\equref{eq: cg noise format bayes}, when using a scale of 1 for \cfgguide,  we essentially sample from a pure conditional model,~\ie,~$p_\theta (\data_{t} \vert \data_{t+1}, \cond)$ on~\equref{eq: guided_diffusion} left side.
The comparison between two plots of ``Scale 1.0'' in~\figref{fig: fractal cg 1.0} and~\figref{fig: fractal cfg} corroborates our analysis in~\secref{sec: cg} that the two sides of~\equref{eq: guided_diffusion} are generally not equal.

\subsection{MNIST}

Our proposed postprocessing step improves the fidelity of generations for both \cgguide~and \cfgguide~on real-world MNIST~\cite{lecun1998gradient} data as shown in~\figref{fig: mnist cg} and~\figref{fig: mnist cfg}.
Based on our analysis in~\secref{sec: postprocess}, this validates our classifier-centric understanding on MNIST.
Our denoising diffusion and rectified flow models are based on a UNet~\cite{Ronneberger2015UNetCN} similar to the one used by~\citet{dhariwal2021diffusion}.
See~\secref{supp sec: mnist exp detail} for experimental details.

\begin{figure*}
    \begin{minipage}[t]{0.2\linewidth}
        \vspace{0pt}
        
\setlength\aboverulesep{0pt}
\captionsetup{width=\columnwidth}
\captionof{table}{
\textbf{Postprocessing for \cfgguide~on CIFAR-10.}
We report conditional FID on 50k generations with seeds from 0 to 49999.
Lower FID is better, and the \colorbox{tabhighlight}{best} in each row is highlighted.
Postprocessing is abbreviated as ``Post.''.
See qualitative results in~\figref{fig:cifar10 edm simplified}.
}
\label{tab: cifar10 edm simplified}
\begin{adjustbox}{width=\linewidth,center}
\setlength{\tabcolsep}{0.1cm}
{
\begin{tabular}{c cc}
\toprule
\multirow{2}{*}{\makecell{CFG Scale\\Before Post.}} & \multicolumn{2}{c}{\makecell{Post.}} \\
\cmidrule(lr){2-3} 
& \xmark & \cmark \\
\hline
2.25 & 8.016 & \cellcolor{tabhighlight}{5.821} \\
2.50 & 9.402 & \cellcolor{tabhighlight}{5.936} \\
2.75 & 10.75 & \cellcolor{tabhighlight}{6.176} \\

\bottomrule
\end{tabular}
}
\end{adjustbox}

    \end{minipage}
    \hfill
    \begin{minipage}[t]{0.79\linewidth}
        \vspace{0pt}
        \centering

        \stepcounter{figure} %
        
        \begin{subfigure}[t]{\linewidth}
            \centering
            \adjincludegraphics[width=\linewidth,trim={{0.0\width} 0 {0\width} 0},clip]{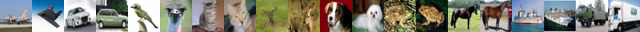}
            \adjincludegraphics[width=\linewidth,trim={{0.0\width} 0 {0\width} 0},clip]{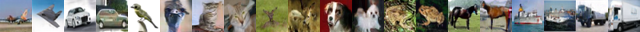}
            \caption{CFG scale before postprocessing is 2.25.}
            \label{subfig:cifar10_edm_simplified_1}
        \end{subfigure}
        \hfill
        \begin{subfigure}[t]{\linewidth}
            \centering
            \adjincludegraphics[width=\linewidth,trim={{0.0\width} 0 {0\width} 0},clip]{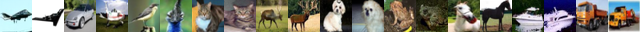}
            \adjincludegraphics[width=\linewidth,trim={{0.0\width} 0 {0\width} 0},clip]{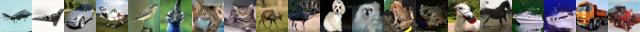}
            \caption{CFG scale before postprocessing is 2.50.}
            \label{subfig:cifar10_edm_simplified_2}
        \end{subfigure}
        \hfill
        \begin{subfigure}[t]{\linewidth}
            \centering
            \adjincludegraphics[width=\linewidth,trim={{0.0\width} 0 {0\width} 0},clip]{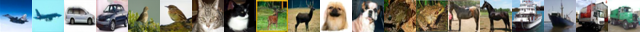}
            \adjincludegraphics[width=\linewidth,trim={{0.0\width} 0 {0\width} 0},clip]{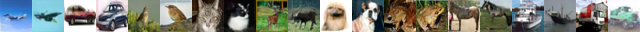}
            \caption{CFG scale before postprocessing is 2.75.}
            \label{subfig:cifar10_edm_simplified_3}
        \end{subfigure}

        \addtocounter{figure}{-1} %
        
        \caption{
        \textbf{\tabref{tab: cifar10 edm simplified}'s visualizations.} In~\figref{subfig:cifar10_edm_simplified_1} to~\ref{subfig:cifar10_edm_simplified_3}, the top and bottom display generations before and after postprocessing.
        As expected, most already high-quality samples remain unchanged.
        } 
        \label{fig:cifar10 edm simplified}
    \end{minipage}
\end{figure*}

\subsection{CIFAR-10}\label{sec: cifar10}

We further verify our classifier-centric understanding on image synthesis via Fr\'echet Inception Distance (FID)~\cite{Heusel2017GANsTB}, a commonly-used metric for generation quality, on  CIFAR-10~\cite{Krizhevsky2009LearningML}.
See~\secref{supp sec: cifar10 exp detail} for details.
We choose EDM~\cite{Karras2022ElucidatingTD}, one of the state-of-the-art diffusion-based generative models on CIFAR-10, as our pre-trained model.
However, the pre-trained EDM model lacks conditioning dropout, making it incompatible with CFG.
We re-train a CFG-compatible EDM, verifying correctness with FID 1.850 (ours with CFG scale 1.0)~\vs~1.849 (pre-trained).
As shown in~\tabref{tab: cifar10 edm simplified}, the postprocessing step improves the generation quality across various guidance scales.
Based on our analysis in~\secref{sec: postprocess}, our classifier-centric understanding holds on this high-dimensional data.

\begin{figure}[!th]
    \vspace{0pt} 
    \centering
    \begin{minipage}[t]{\columnwidth}
        \begin{subfigure}[t]{0.49\textwidth}
            \centering
            \adjincludegraphics[width=\textwidth,trim={{0.0\width} 0 {0\width} 0},clip]{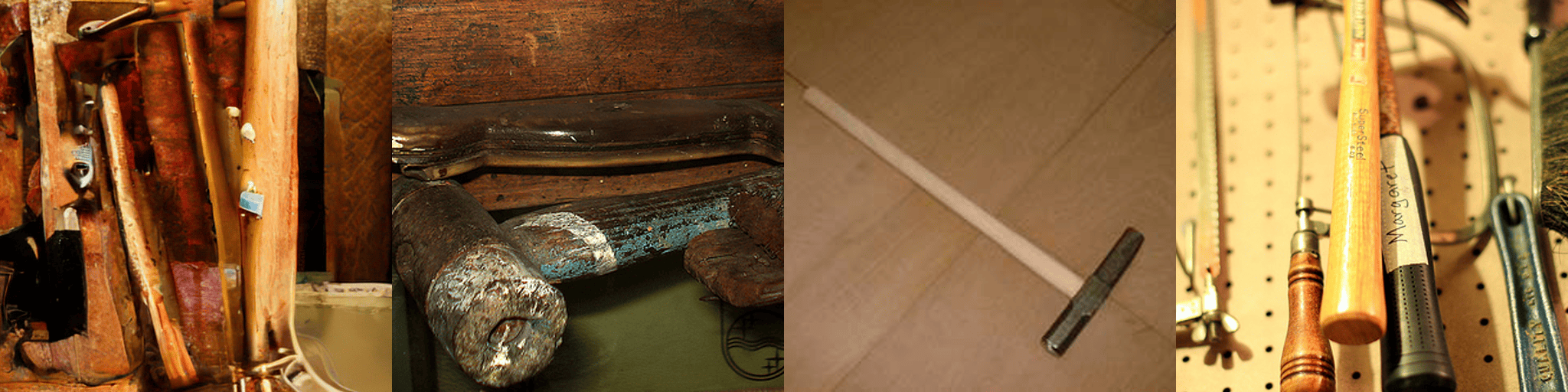}
            \captionsetup{width=\textwidth}
            \caption{Class 587 (hammer).}
            \label{fig: imagenet nn compare 1}
        \end{subfigure}
        \hfill
        \begin{subfigure}[t]{0.49\textwidth}
            \centering
            \adjincludegraphics[width=\textwidth,trim={{0.0\width} 0 {0\width} 0},clip]{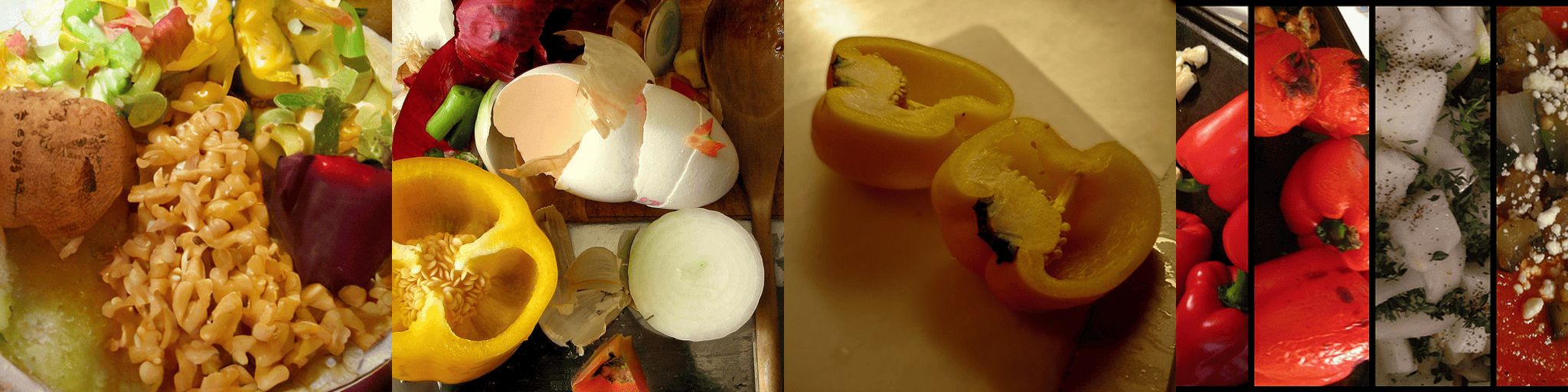}
            \captionsetup{width=\textwidth}
            \caption{Class 945 (bell pepper).}
            \label{fig: imagenet nn compare 2}
        \end{subfigure}
    \end{minipage}%
    \hfill
    \begin{minipage}[t]{\columnwidth}
        \begin{subfigure}[t]{0.49\textwidth}
            \centering
            \adjincludegraphics[width=\textwidth,trim={{0.0\width} 0 {0\width} 0},clip]{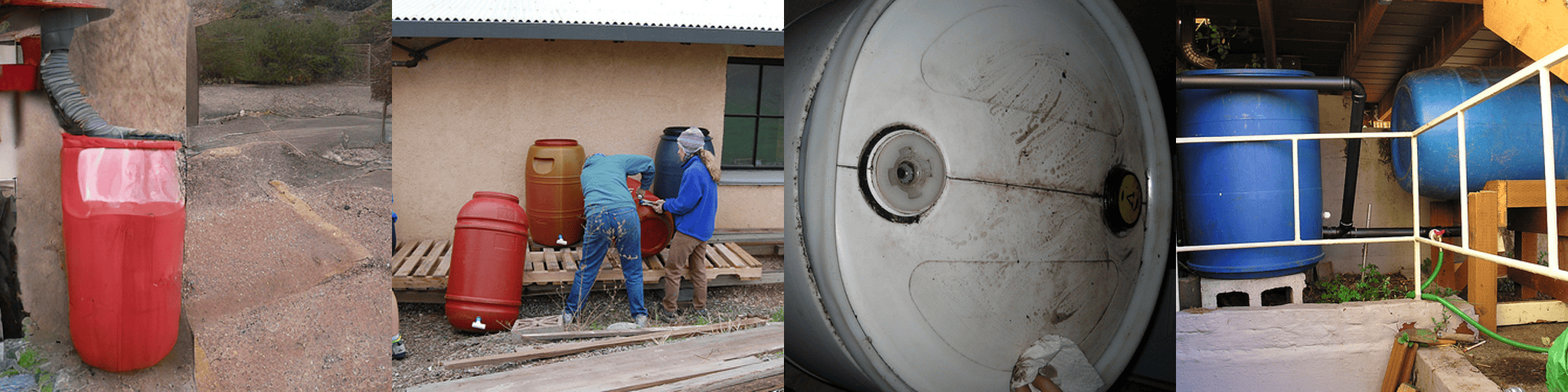}
            \captionsetup{width=\textwidth}
            \caption{Class 756 (rain barrel).}
            \label{fig: imagenet nn compare 3}
        \end{subfigure}
        \hfill
        \begin{subfigure}[t]{0.49\textwidth}
            \centering
            \adjincludegraphics[width=\textwidth,trim={{0.0\width} 0 {0\width} 0},clip]{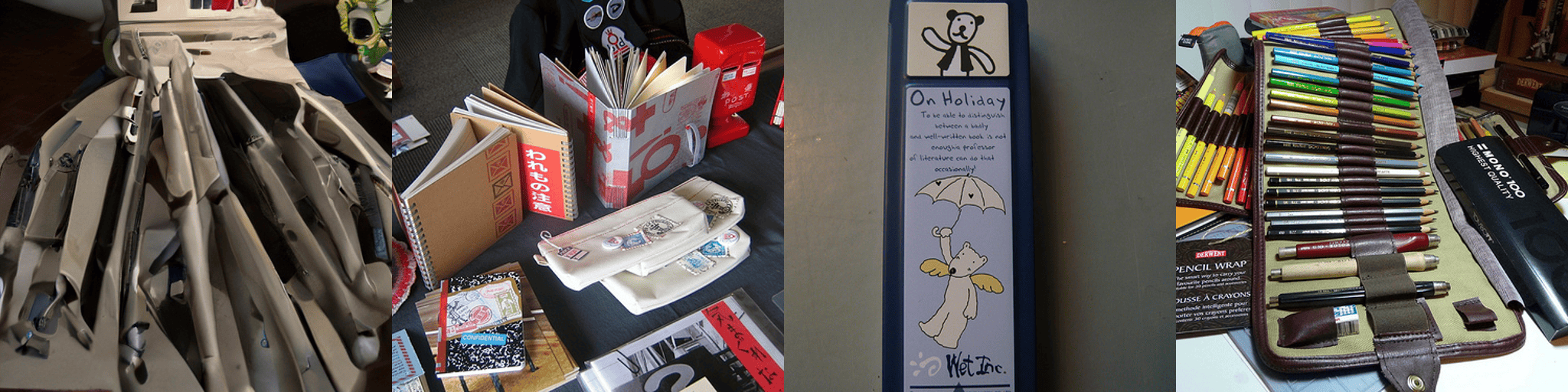}
            \captionsetup{width=\textwidth}
            \caption{Class 709 (pencil box).}
            \label{fig: imagenet nn compare 4}
        \end{subfigure}
    \end{minipage}%
    \caption{
      \textbf{Nearest neighbor based on different distance metrics for ImageNet $512^2$.}
      For each plot, from left to right, we display the generation from pre-traiend EDM2~\cite{Karras2023AnalyzingAI} and its nearest neighbor from the real images based on Euclidean distance in the feature space of DINOv2~\cite{oquab2023dinov2}, Stable Diffusion's VAE~\cite{Rombach2021HighResolutionIS},
      and state-of-the-art classifier from \texttt{timm}~\cite{rw2019timm} respectively.
      Different distance metrics can produce significantly different results.
    }
    \label{fig: imagenet nn compare}
\end{figure}

\section{Discussion}\label{sec: discussion}

\noindent\textbf{\texttt{NN} in~\equref{eq: post flow}.}
Unlike the 2D Fractal case in~\secref{sec: 2d fractal}, where Euclidean distance is well-defined, there’s no clear distance definition for high-dimensional data. In an ablation on CIFAR-10 in~\secref{supp sec: nn choice}, we find that while our postprocessing consistently improves overall generation quality,~\ie,~it can be a proxy for assessing our classifier-centric understanding (\secref{sec: postprocess}), specific performance can vary.

\noindent\textbf{On ImageNet~\cite{deng2009imagenet}.}
We experimented with our postprocessing on ImageNet $512^2$ with various distance metrics, but the performance is not satisfactory. Due to the intricate structure of this high-dimensional space, determining a reasonable distance metric to apply~\equref{eq: post flow} is difficult, and this is an active research area~\cite{Aggarwal2001OnTS, Qian2014FinegrainedVC, Zhang2018TheUE, Stein2023ExposingFO}.
In~\figref{fig: imagenet nn compare}, we list some qualitative examples on nearest neighbors with distance space we tried, spanning features from self-supervised model~\cite{oquab2023dinov2}, VAE~\cite{rombach2022high}, and classifier~\cite{rw2019timm}.
The behavior differs significantly.
Thus, we leave the experiment as a future work.

\noindent\textbf{Limitations.}
Since our postprocessing step runs another round of diffusion, to verify our classifier-centric understanding, inference time will be doubled when compared to the generation process without postprocessing.
However, with more prevailing distillation techniques and faster samplers, we think the overhead can be largely mitigated.

\section{Conclusion}\label{sec: conclusion}

We carry out an empirical study aiming to understand \cfgguide~from a  classifier-centric perspective.
Our analysis
reveals that both \cfgguide~and \cgguide~push the denoising diffusion process away from the data's decision boundaries on 1D data.
For high-dimensional data, we propose a flow matching based postprocessing step to indirectly assess our classifier-centric understanding and verify its effectiveness across datasets.

\section*{Acknowledgments}
Work supported in part by NSF grants 2008387, 2045586, 2106825, MRI 1725729, and NIFA award 2020-67021-32799.

\bibliography{aaai2026}

\beginsupplement
\appendix

\twocolumn[\section*{Appendix -- \ourtitle}]

\noindent{}This supplementary material is structured as follows:
\begin{enumerate}
    \item \secref{supp sec: proof} derives the formulation for \cgguide, which is copied from~\cite{dhariwal2021diffusion} for completeness purposes;
    \item \secref{supp sec: generalized} discusses a generalized postprocessing model and its evaluation on 2D fractal and MNIST data;
    \item \secref{supp sec: nn choice} discusses the choice of nearest neighbor in high-dimensional data;
    \item \secref{sec: implement} describes implementation details;
    \item \secref{supp sec: broader impact} discusses the broader impact;
    \item \secref{sec: more vis} provides more visualizations.
\end{enumerate}

\section{Derivation for~\equref{eq: guided_diffusion}}\label{supp sec: proof}

For completeness, we copy the derivation from Appendix H of~\citet{dhariwal2021diffusion}, with only minimal adjustments to align it with our notation.
Throughout the derivation, we \textcolor{orange}{highlight} the steps that rely on the key assumption we mentioned in~\equref{eq: guided_diffusion_key_step}, demonstrating the validity of our question raised in~\secref{sec: cg},~\ie,~whether the conditional diffusion process decomposition in~\equref{eq: guided_diffusion} always holds.

Specifically,~\citet{dhariwal2021diffusion} first define a conditional Markovian process $\hat{q}$ as
\begin{align}
    \hat{q}(\data_0) &\triangleq q(\data_0) \\
    \hat{q}(c|\data_0) &\triangleq \text{Known labels per sample} \\
    \color{orange} \hat{q}(\data_{t+1}|\data_t,c) &\color{orange} \triangleq q(\data_{t+1} | \data_t) \quad \text{(a key assumption) }\label{supp eq: key assumption} \\
    \hat{q}(\data_{1:T}|\data_0,c) &\triangleq \prod_{t=1}^T \hat{q}(\data_t|\data_{t-1},c)  \quad \text{(Markovian assumption)} \label{supp eq: markov}
\end{align}
\equref{supp eq: key assumption} indicates that the defined process $\hat{q}$ behaves identical to an unconditional process $q$.~\citet{dhariwal2021diffusion} then prove the following property for the newly-defined process $\hat{q}$:
\begin{align}
    \hat{q}(\data_{t+1}|\data_t) &= \int_{c} \hat{q}(\data_{t+1},c|\data_t) \,dy \\
                   &= \int_{c} \hat{q}(\data_{t+1}|\data_t,c) \, \hat{q}(c|\data_t) \,dc \\
                   &\color{orange} = \int_{c} q(\data_{t+1}|\data_t) \, \hat{q}(c|\data_t) \,dc \quad \text{(due to~\equref{supp eq: key assumption})} \\
                   &= q(\data_{t+1}|\data_t) \int_{c} \hat{q}(c|\data_t) \,dc \\
                   &= q(\data_{t+1}|\data_t) \label{supp eq: hat_q x_t+1 x_t same as q} \\
                   &\color{orange} = \hat{q}(\data_{t+1}|\data_t,c) \quad \text{(due to~\equref{supp eq: key assumption})} \label{supp eq: hat_q same wo cond}
\end{align}

Following a similar logic,~\citet{dhariwal2021diffusion} further derive $\hat{q}(\data_{1:T}|\data_0)$
\allowdisplaybreaks  %
\begin{align}
    &= \int_{c} \hat{q}(\data_{1:T},c|\data_0) \,dc \\
    &= \int_{c} \hat{q}(c|\data_0) \, \hat{q}(\data_{1:T}|\data_0,c) \,dc \\
    &= \int_{c} \hat{q}(c|\data_0) \prod_{t=1}^T \hat{q}(\data_t|\data_{t-1},c) \,dc \quad \text{(due to~\equref{supp eq: markov})} \\
    &\color{orange} = \int_{c} \hat{q}(c|\data_0) \prod_{t=1}^T q(\data_t|\data_{t-1}) \,dc \quad \text{(due to~\equref{supp eq: key assumption})} \\
    &= \prod_{t=1}^T q(\data_t|\data_{t-1}) \int_{c} \hat{q}(c|\data_0) \,dc \\
    &= \prod_{t=1}^T q(\data_t|\data_{t-1}) \\
    &= q(\data_{1:T}|\data_0) \label{supp eq: cond joint dist}
\end{align}

With~\equref{supp eq: cond joint dist} in hand,~\citet{dhariwal2021diffusion} derive $\hat{q}(\data_t)$
\allowdisplaybreaks  %
\begin{align}
    &= \int_{\data_{0:t-1}} \hat{q}(\data_0, ..., \data_t) \,d\data_{0:t-1} \\
    &= \int_{\data_{0:t-1}} \, \hat{q}(\data_0) \hat{q}(\data_1,...,\data_t|\data_0) \,d\data_{0:t-1} \\
    &\color{orange} = \int_{\data_{0:t-1}} \, q(\data_0) q(\data_1,...,\data_t|\data_0) \,d\data_{0:t-1} \; \text{(due to~\equref{supp eq: cond joint dist})}\\
    &= \int_{\data_{0:t-1}} q(\data_0,...,\data_t) \,d\data_{0:t-1} \\
    &= q(\data_t) \label{supp eq: hat q x_t same as q}
\end{align}

Using the identities in~\equref{supp eq: hat q x_t same as q} and $\hat q(\data_{t+1}|\data_t) = q(\data_{t+1}|\data_t)$ in~\equref{supp eq: hat_q x_t+1 x_t same as q}, it is trivial to show via Bayes rule that the unconditional reverse process $\hat{q}(\data_t|\data_{t+1}) = q(\data_t|\data_{t+1})$.
Another observation about $\hat{q}$ is that it gives rise to a noisy classification function, $\hat{q}(c|\data_t)$.
\citet{dhariwal2021diffusion} show that this classification distribution does not depend on $\data_{t+1}$ (a noisier version of $\data_t$):
\begin{align}
    \hat{q}(c|\data_t,\data_{t+1})
    &= \hat{q}(\data_{t+1}|\data_t,c) \, \frac{\hat{q}(c|\data_t)}{\hat{q}(\data_{t+1}|\data_t)} \\
    &\color{orange} = \hat{q}(\data_{t+1}|\data_t) \, \frac{\hat{q}(c|\data_t)}{\hat{q}(\data_{t+1}|\data_t)} \quad \text{(due to~\equref{supp eq: hat_q same wo cond})} \\
    &= \hat{q}(c|\data_t) \label{supp eq: classifier no x_t+1}
\end{align}

\citet{dhariwal2021diffusion} finally derive the conditional reverse process for $\hat{q}$ as
\begin{align}
    \hat{q}(\data_t|\data_{t+1},c)
    &= \frac{\hat{q}(\data_t,\data_{t+1},c)}{\hat{q}(\data_{t+1},c)} \\
    &= \frac{\hat{q}(\data_t,\data_{t+1},c)}{\hat{q}(c|\data_{t+1}) \, \hat{q}(\data_{t+1})} \\
    &= \frac{\hat{q}(\data_t|\data_{t+1}) \, \hat{q}(c|\data_t,\data_{t+1}) \, \hat{q}(\data_{t+1})}{\hat{q}(c|\data_{t+1}) \, \hat{q}(\data_{t+1})} \\
    &= \frac{\hat{q}(\data_t|\data_{t+1}) \, \hat{q}(c|\data_t,\data_{t+1})}{\hat{q}(c|\data_{t+1})} \\
    &\color{orange} = \frac{\hat{q}(\data_t|\data_{t+1}) \, \hat{q}(c|\data_t)}{\hat{q}(c|\data_{t+1})} \quad \text{(due to~\equref{supp eq: classifier no x_t+1})} \\
    &\color{orange} = \frac{q(\data_t|\data_{t+1}) \, \hat{q}(c|\data_t)}{\hat{q}(c|\data_{t+1})} \quad \text{(due to~\equref{supp eq: hat q x_t same as q})} \label{supp eq: guided_diffusion}
\end{align}

The $\hat{q}(c|\data_{t+1})$ term in the denominator can be treated as a constant since it does not depend on $\data_t$.
\citet{dhariwal2021diffusion} thus sample from the distribution $Z q(\data_t|\data_{t+1})\hat{q}(c|\data_t)$ where $Z$ is a normalizing constant.
\equref{supp eq: guided_diffusion} is essentially~\equref{eq: guided_diffusion}, the foundation for both \cgguide~and~\cfgguide~works.

As can be seen from the derivation,~\equref{supp eq: key assumption},~\ie,~\equref{eq: guided_diffusion_key_step}, is the most important assumption.
Thus our analysis in~\secref{sec: analysis} is valid.

\section{One-for-All-Scales Postprocessing Model}\label{supp sec: generalized}

In the main paper, specifically in \figref{fig: fractal cg},~\ref{fig: fractal cfg}, and~\ref{fig: mnist cg}, we show results using a dedicated flow matching model which was trained on a set of samples $\set$ generated with a \textit{specific} guidance scale.
A natural next question: is it possible to develop a \textit{generalized} postprocessing model that can be trained only once and then applied to all samples regardless of their guidance scale?

\begin{figure*}[!t]
    \vspace{0pt} 
    \centering
    \begin{minipage}[t]{\textwidth}
        \begin{subfigure}[t]{\textwidth}
            \centering
            \adjincludegraphics[width=\textwidth,trim={{0.0\width} 0 {0\width} 0},clip]{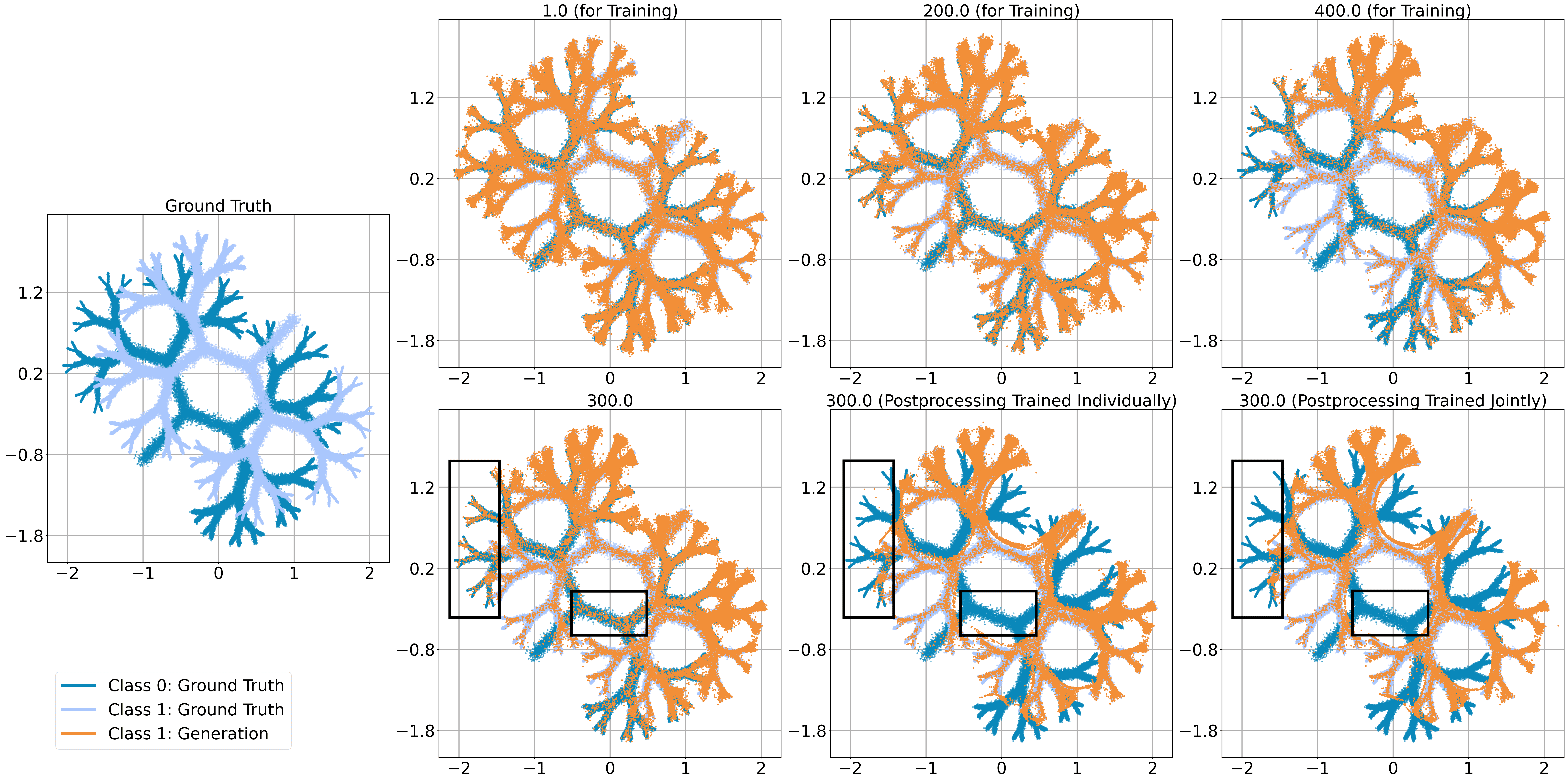}
            \captionsetup{width=\textwidth}
            \caption{
              \textbf{Performance comparison on data with \cgguide.}
            }
            \label{fig: fractal cascade compare cg}
        \end{subfigure}
    \end{minipage}%
    \hfill
    \begin{minipage}[t]{\textwidth}
        \begin{subfigure}[t]{\textwidth}
            \centering
            \adjincludegraphics[width=\textwidth,trim={{0.0\width} 0 {0\width} 0},clip]{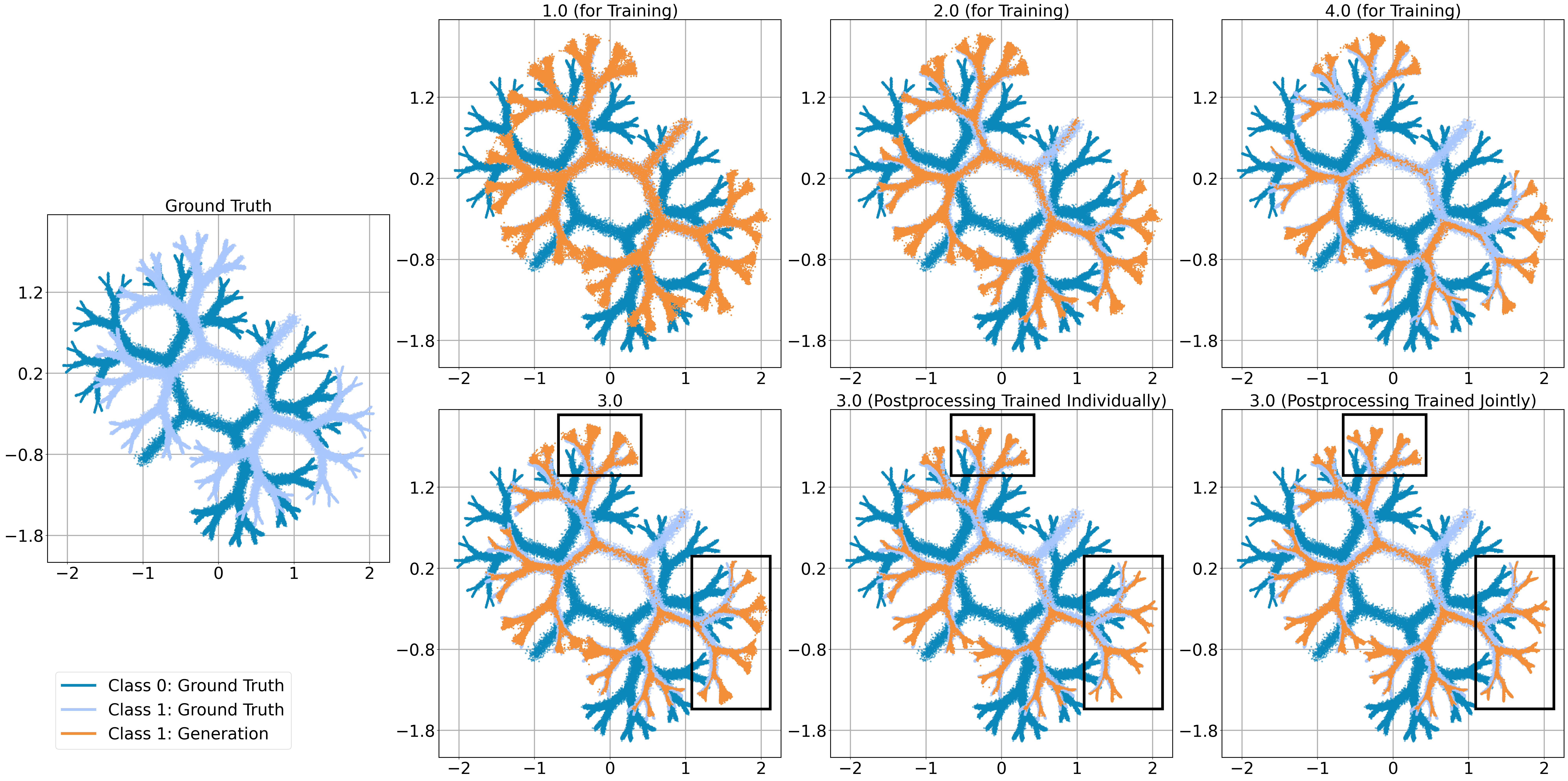}
            \captionsetup{width=\textwidth}
            \caption{
              \textbf{Performance comparison on data with \cfgguide.}
            }
            \label{fig: fractal cascade compare cfg}
        \end{subfigure}
    \end{minipage}%
    \caption{
      \textbf{One-for-All-Scales postprocessing model  (\secref{sec: postprocess}) on 2D fractal data.}
      We train a \textit{generalized} postprocessing model that can be applied to samples generated by various guidance scales.
      In~\figref{fig: fractal cascade compare cg} and~\ref{fig: fractal cascade compare cfg}, the top row displays the training data set composed of samples generated with three different scales.
      Note that for clear visualization, we only show one class of data. The generalized model is trained on two classes of samples.
      The bottom row in~\figref{fig: fractal cascade compare cg} and~\ref{fig: fractal cascade compare cfg} compare the performance between \textit{individually-trained} and \textit{generalized} postprocessing model.
      We want to emphasize that \textit{we keep the amount of training data identical for both post-processing models such that the comparison is fair.}
      Concretely, for the individually-trained model, we generate 100k points with the specific guidance scale. For the generalized model, we generate 100k / $K$ samples for each of the $K$ different guidance scales.
      As can be seen clearly, the generalized postprocessing model performs well, demonstrating the effect of a One-for-All-Scales postprocessing model: it can ease the analysis of our classifier-centric understanding.
    }
    \label{fig: fractal cascade compare}
\end{figure*}

\begin{figure*}[!t]
    \vspace{0pt} 
    \centering
    \begin{minipage}[t]{\textwidth}
        \begin{subfigure}[t]{0.32\textwidth}
            \centering
            \adjincludegraphics[width=\textwidth,trim={{0.0\width} 0 {0\width} 0},clip]{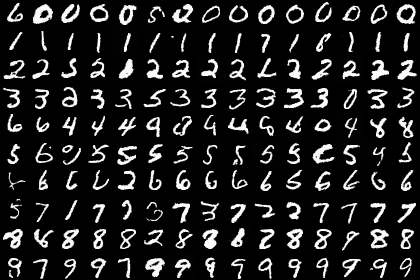}
            \captionsetup{width=\textwidth}
            \caption{
              \textbf{Before postprocessing.}
            }
            \label{fig: mnist cascade compare cg before}
        \end{subfigure}
        \hfill
        \begin{subfigure}[t]{0.32\textwidth}
            \centering
            \adjincludegraphics[width=\textwidth,trim={{0.0\width} 0 {0\width} 0},clip]{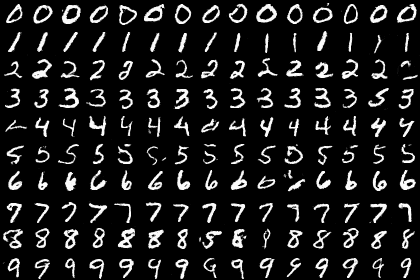}
            \captionsetup{width=\textwidth}
            \caption{
              \textbf{Postprocessing; individually-trained.}
            }
            \label{fig: mnist cascade compare cg after indiv}
        \end{subfigure}
        \hfill
        \begin{subfigure}[t]{0.32\textwidth}
            \centering
            \adjincludegraphics[width=\textwidth,trim={{0.0\width} 0 {0\width} 0},clip]{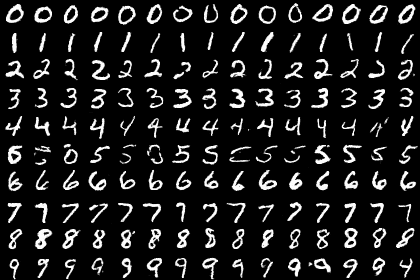}
            \captionsetup{width=\textwidth}
            \caption{
              \textbf{Postprocessing; jointly-trained.}
            }
            \label{fig: mnist cascade compare cg after joint}
        \end{subfigure}
    \end{minipage}%
    \hfill
    \begin{minipage}[t]{\textwidth}
        \begin{subfigure}[t]{\textwidth}
            \centering
            \adjincludegraphics[width=\textwidth,trim={{0.0\width} 0 {0\width} 0},clip]{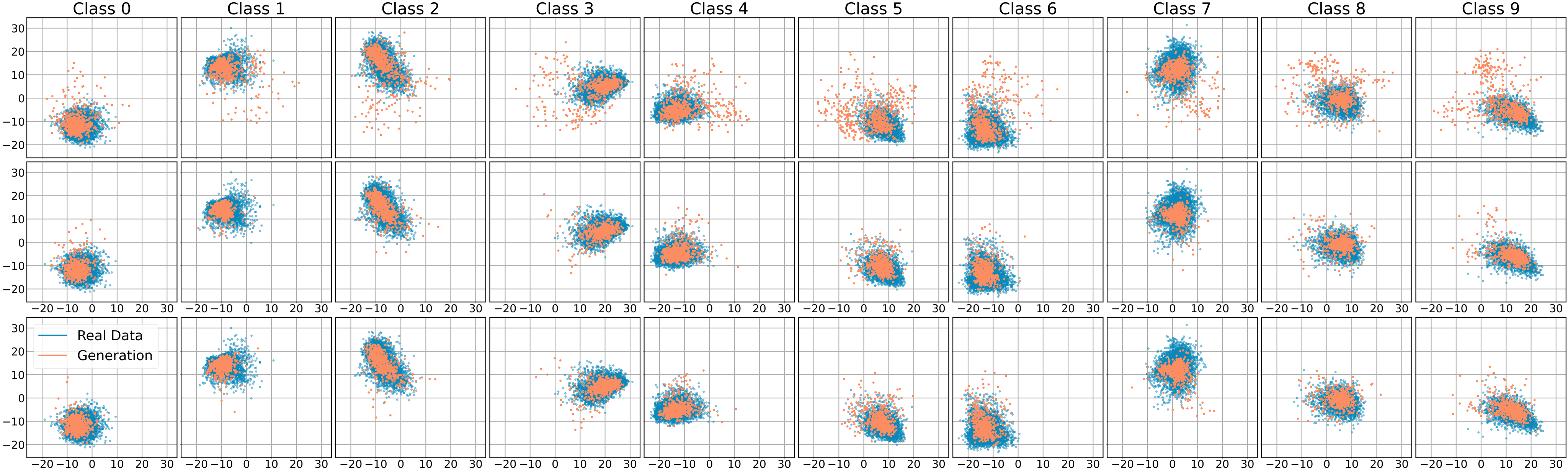}
            \captionsetup{width=\textwidth}
            \caption{
              \textbf{PCA comparison for generalized postprocessing model.}
              We show PCA results with a procedure similar to that in~\figref{fig: mnist cg pca}.
              The $1^\text{st}$, $2^\text{nd}$, and $3^\text{rd}$ rows correspond to~\figref{fig: mnist cascade compare cg before},~\ref{fig: mnist cascade compare cg after indiv}, and~\ref{fig: mnist cascade compare cg after joint} respectively.
              The postprocessing moves the learned distribution (orange clusters) closer to the real one (blue clusters).
              The generalized model performs on par with the model trained with generations from the specific guidance scale.
            }
            \label{fig: mnist cascade compare cg pca}
        \end{subfigure}
    \end{minipage}%
    \caption{
      \textbf{One-for-All-Scales postprocessing model  (\secref{sec: postprocess}) for \cgguide~on MNIST.}
      Here we show the performance comparison for different postprocessing models on generations with a scale of 3.0.
      \figref{fig: mnist cascade compare cg before},~\ref{fig: mnist cascade compare cg after indiv}, and~\ref{fig: mnist cascade compare cg after joint} share the same initial noises for corresponding cells and conditioning information from top to bottom row is the digit 0 to 9.
      We want to emphasize that \textit{we keep the amount of training data identical for both post-processing models such that the comparison is fair.}
      Concretely, for the individually-trained model, we generate 6k images with the specific guidance scale of 3.0. For the generalized model, we generate 2k samples each with scales 1.0, 5.0, and 10.0 respectively.
      As can be seen clearly, the generalized model performs well even though it has not been exposed to samples generated with the guidance scale of 3.0 (also verified by~\figref{fig: mnist cascade compare cg pca}).
    }
    \label{fig: mnist cascade compare cg}
\end{figure*}

\begin{figure*}[!t]
    \vspace{0pt} 
    \centering
    \begin{minipage}[t]{\textwidth}
        \begin{subfigure}[t]{0.32\textwidth}
            \centering
            \adjincludegraphics[width=\textwidth,trim={{0.0\width} 0 {0\width} 0},clip]{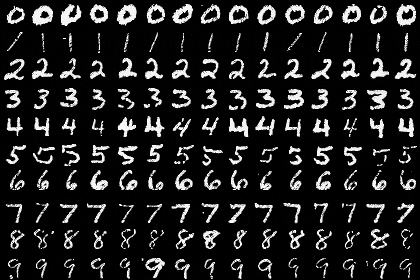}
            \captionsetup{width=\textwidth}
            \caption{
              \textbf{Before postprocessing.}
            }
            \label{fig: mnist cascade compare cfg before}
        \end{subfigure}
        \hfill
        \begin{subfigure}[t]{0.32\textwidth}
            \centering
            \adjincludegraphics[width=\textwidth,trim={{0.0\width} 0 {0\width} 0},clip]{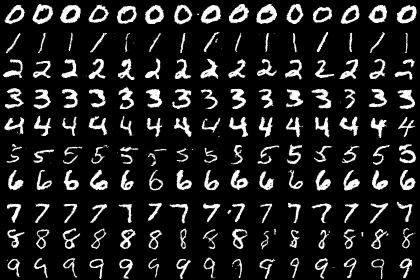}
            \captionsetup{width=\textwidth}
            \caption{
              \textbf{Postprocessing; individually-trained model.}
            }
            \label{fig: mnist cascade compare cfg after indiv}
        \end{subfigure}
        \hfill
        \begin{subfigure}[t]{0.32\textwidth}
            \centering
            \adjincludegraphics[width=\textwidth,trim={{0.0\width} 0 {0\width} 0},clip]{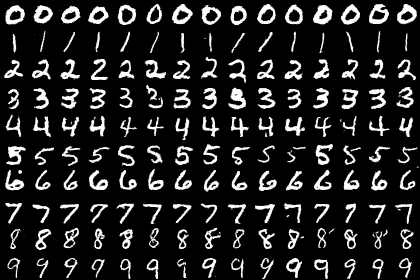}
            \captionsetup{width=\textwidth}
            \caption{
              \textbf{Postprocessing; jointly-trained model.}
            }
            \label{fig: mnist cascade compare cfg after joint}
        \end{subfigure}
    \end{minipage}%
    \caption{
      \textbf{One-for-All-Scales postprocessing model  (\secref{sec: postprocess}) for \cfgguide~on MNIST.}
      Here we show the performance comparison for different postprocessing models on generations with a scale of 8.0.
      \figref{fig: mnist cascade compare cfg before},~\ref{fig: mnist cascade compare cfg after indiv}, and~\ref{fig: mnist cascade compare cfg after joint} share the same initial noises for corresponding cells and conditioning information from top to bottom row is the digit 0 to 9.
      We want to emphasize that \textit{we keep the amount of training data identical for both post-processing models such that the comparison is fair.}
      Concretely, for the individually-trained model, we generate 6k images with the specific guidance scale of 8.0. For the generalized model, we generate 2k samples each with scales 1.0, 5.0, and 10.0, respectively.
      Both individually-trained and generalized postprocessing models recover the fidelity, verifying our classifier-centric understanding.
    }
    \label{fig: mnist cascade compare cfg}
\end{figure*}

We conduct experiments and answer this question in the affirmative.
Specifically, we show results for 2D fractal data in~\figref{fig: fractal cascade compare}, while~\figref{fig: mnist cascade compare cg} and~\figref{fig: mnist cascade compare cfg} provide results for MNIST.
We emphasize that the amount of training data is identical,~\ie,~the generalized model is exposed to the same amount of training data as the individually-trained model. 
From the visualized results we observe that even in this fair setup, the generalized model performs on-par or even better than individually-trained models.
We hypothesize that this is because generations with different scales essentially share a similar underlying distribution, as  \cgguide~and \cfgguide~both  push the denoising trajectories away from the decision boundaries.
With a combined set of samples, the model is exposed to a  diverse but coherent dataset, facilitating model generalization.
This one-for-all-scales model eases our analysis on high-dimensional data, as it is costly to train separate postprocessing models for each guidance scale.
Specifically, for the experiments on CIFAR-10 in~\secref{sec: cifar10} and~\secref{supp sec: nn choice}, the results are reported from one-for-all-scales models.

\begin{table}[!t]
\setlength\aboverulesep{0pt}
\captionsetup{width=\linewidth}
\caption{
\textbf{One-for-All-Scales postprocessing model for \cfgguide~on CIFAR-10 (base model as flow matching).}
We report conditional FID evaluation computed on 50k generations (lower is better).
We train a generalized flow matching based postprocessing model on data sampled with guidance scales 1.00, 2.00, and 3.00 and apply it to samples with scales that are not seen during training.
We ablate the distance space where nearest neighbor $\texttt{NN}$ is computed, namely 1) the raw RGB space (Pixel); 2) the feature for the \texttt{CLS} token output by DINOv2~\cite{oquab2023dinov2} (DINOv2 \texttt{CLS}); and 3) the average of the patch features (DINOv2 Patch).
We also ablate the number of top-$k$ nearest neighbors (\texttt{NN}) used during training of the postprocessing model.
Please refer to~\secref{supp sec: nn choice} for details.
Postprocessing is abbreviated as ``Post.'' and \colorbox{tabhighlight}{best FID} is highlighted.
}
\label{tab: cifar10}
\begin{adjustbox}{width=0.8\linewidth,center}
\setlength{\tabcolsep}{0.15cm}
{
\begin{tabular}{lcllrrr}
\toprule
& \multirow{2}{*}{\makecell{Post.}} & \multirow{2}{*}{\makecell{\texttt{NN} Space}} & \multirow{2}{*}{\makecell{top-$k$ \\for \texttt{NN}}} & \multicolumn{3}{c}{\footnotesize CFG Scale Before Post.} \\
\cmidrule(lr){5-7}
&  &  &  & 2.25 & 2.50 & 2.75 \\
\hline
1 & \xmark & -- & -- & 35.77 & 41.58 & 46.37  \\
\hline
2 & \cmark &  Pixel & 20 & 22.55 & 25.96 & 28.95   \\
3 & \cmark & DINOv2 \texttt{CLS} & 20 & 19.37 & 22.97 & 26.48   \\
4 & \cmark & DINOv2 Patch & 20 & \cellcolor{tabhighlight}{17.27} & 20.19 & 23.32   \\
\hline
5 & \cmark & DINOv2 Patch & 40 & 17.48 & \cellcolor{tabhighlight}{20.10} & \cellcolor{tabhighlight}{22.69} \\
\bottomrule
\end{tabular}
}
\end{adjustbox}
\end{table}

\section{Nearest Neighbor in High-Dimensional Data: Case Study on CIFAR-10}\label{supp sec: nn choice}

As briefly discussed in~\secref{sec: discussion}, 
the space for nearest neighbor computation,~\ie,~\texttt{NN} in~\equref{eq: post flow}, is an important design choice.
However, unlike 2D Fractal data, where Euclidean distance is well-defined, there is no golden choice for the distance on high-dimensional data.
To understand the effects of different distance metrics, we carry out a case study on CIFAR-10.
Further, we are also interested in knowing whether our classifier-centric understanding is effective for flow-matching-based base generation model.
Thus, we study how our proposed postprocessing step works in conjunction with a base generative model that is based on flow matching,~\ie,~both our base and postprocessing generative models are flow-matching-based models.

For this, we first train a rectified flow~\cite{Liu2022FlowSA} based conditional generative model on CIFAR-10.
Then, following the procedure for a one-for-all-scales postprocessing model discussed in Appendix~\ref{supp sec: generalized}, we compose a training set of 50k generations sampled with a guidance scale of 1.0, 2.0, and 3.0. 
Each scale contributes to one-third of the training data.
With this data, we train a generalized postprocessing model and evaluate it using Fr\'echet Inception Distance (FID)~\cite{Heusel2017GANsTB} computed with generations with unseen guidance scales.
All generations are obtained using the adaptive solver Dopri5 following the suggestions of~\citet{Tong2023ImprovingAG}.

Here we study two possible candidates for the \texttt{NN} space:
1) directly using the raw RGB pixel;
2) using the feature embedding space of recently developed  pre-trained foundation models.
For the latter, we employ  DINOv2~\cite{oquab2023dinov2} pre-trained with registers~\cite{darcet2023vitneedreg}.
Due to the ViT structure~\cite{Dosovitskiy2020AnII} used in DINOv2, we can either use the information encoded in the \texttt{CLS} token to search for the nearest neighbor or we can take an average of the features corresponding to each patch token.
The quantitative results are summarized in~\tabref{tab: cifar10}.
We observe that different distance metric indeed produces drastically different performance.
However, in the CIFAR-10 case, the postprocessing model consistently holds the trend of improving the overall generation quality.
This verifies our classifier-centric understanding on CIFAR-10.

\section{Experiment Details}\label{sec: implement}

\subsection{Top-$k$ Nearest Neighbor (\texttt{NN}) Search}

There are several great tools~\cite{soar_2023, avq_2020} for accelerating the nearest neighbor (\texttt{NN}) search required in~\equref{eq: post flow}.
In this work, we use FAISS~\cite{douze2024faiss} for all our top-$k$ nearest neighbor computations.

\subsection{Hardware Setup}\label{supp sec: compute setup}

Due to limited computing resources, our experiments were conducted on a mix of available hardware, including NVIDIA A6000, L40S, and A100 GPUs.
On an A100 80GB GPU, our postprocessing uses 1.6 GB more GPU memory and introduces an overhead of 4.5 seconds for each image generation in the experiment on CIFAR-10 in~\secref{sec: cifar10}.

\subsection{Experiments on 1D Gaussian}

\subsubsection{Models}\label{sec: model 1d gaussian}

\begin{figure*}[!t]
    \centering
    \includegraphics[width=\textwidth]{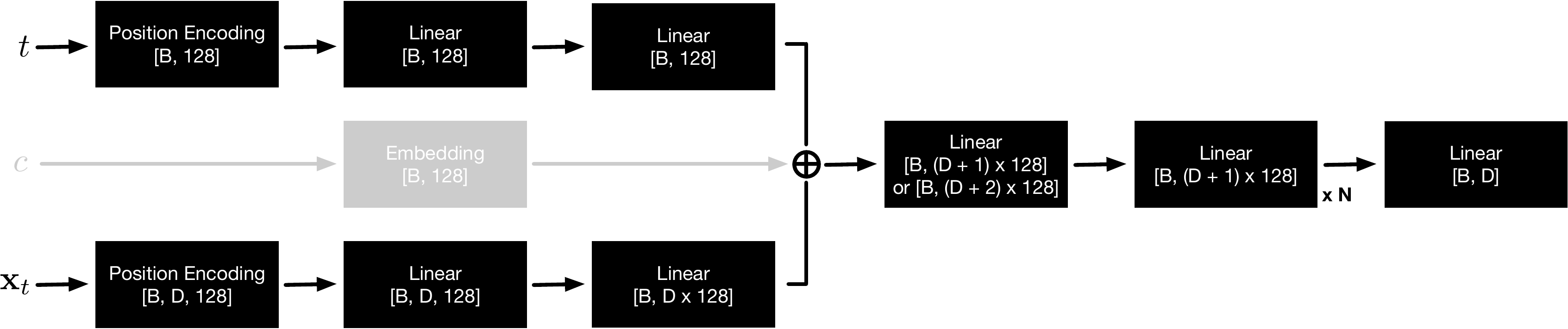}
    \caption{
      \textbf{MLP-based denoising diffusion model.} 
      Here we provide details about the network structure for our MLP-based denoising diffusion model, which is used for experiments on 1D Gaussian data in~\secref{sec: analysis} and on 2D fractal data in~\secref{sec: 2d fractal}.
      In each block, the top row provides the layer's name. The bottom row illustrates the tensor shape after the corresponding layer's processing. Here, $B$ and $D$ stand for batch size and data channel respectively.
      For example, $D$ equals 1 and 2 respectively for the data of 1D Gaussian and 2D fractal.
      We have the following position encoding function applied on \textit{each channel} of the input data: $\{\sin(2\pi \cdot u_0),\dots,\sin(2\pi \cdot u_{C-1}), \cos(2\pi \cdot u_0),\dots,\cos(2\pi \cdot u_{C-1})\}$, where $C=128 / 2$ in our setup.
      We use $u_i = x \cdot \exp\left( i \cdot  \log_2\left( -\frac{10000}{C - 1}\right)\right)$, where $x$ is the value at the corresponding channel the position encoding is applied.
      Blocks with a grey background color are optional.
      Concretely, for \cgguide, we train an \textit{unconditional} model, which does not use the conditioning information $c$. In contrast, for \cfgguide, we enable the embedding layer to train a \textit{conditional} model.
      Throughout our experiments, we repeat the second to last block four times, namely we have $N=4$.
    }
    \label{fig: mlp point diffusion}
\end{figure*}

For 1D Gaussian data, both our denoising diffusion model for \cgguide~and \cfgguide~are based on the structure illustrated in~\figref{fig: mlp point diffusion}.
We also construct the nonlinear classifier for the \cgguide~results shown in~\figref{fig: gaussian traj cg classifier nonlinear} using the structure illustrated in~\figref{fig: mlp point diffusion}, but remove the class embedding block.
The linear classifier used for the results shown in~\figref{fig: gaussian traj cg classifier linear} is developed by feeding the concatenated data $\data_t$ and diffusion step $t$ to a linear layer.

We train both 1) linear and nonlinear classifiers; and 2) unconditional and conditional diffusion models using the optimizer AdamW~\cite{Loshchilov2017DecoupledWD} with a batch size of 4096 points and a learning rate of $10^{-4}$.
The classifiers and diffusion models are trained for 50k and 100k steps, respectively.
We run each evaluation once.

\subsection{Experiments on 2D Fractal}

\subsubsection{2D Fractal Construction}\label{sec: fractal construct}

The fractal data,~\eg,~the data presented in~\figref{fig: fractal cg} and~\figref{fig: fractal cfg}, is synthesized similarly to the one used by \citet{Karras2024GuidingAD} (Appendix~C in~\cite{Karras2024GuidingAD}).
The fractal data consists of 2D points sampled from a mixture of Gaussians, each situated to form branches.

Concretely, we first define how to conduct a branch split.
For each class, the corresponding fractal is essentially a binary tree.
The $i$-th branch is defined by the starting position $\mathbf{s}_i \in \mathbb{R}^2$, length $l_i \in \mathbb{R}$, and orientation $o_i \in [0, 2 \pi]$.
To obtain the child branch,~\eg,~the $k$-th branch where $k\in\{2i, 2i + 1\}$, we follow the procedure:
\begin{align}
    \mathbf{s}_{k} &= \mathbf{s}_i + l_i \cdot \left( \cos(o_i), \sin(o_i) \right)^T, \\
    l_{k} &= l_i \cdot (1 - 0.4 \cdot \xi),\; \xi \sim U(0.5, 0.8), \quad\text{and} \\
    o_{k} &= o_i + (-1)^{k + 1} \cdot \pi \cdot \left( \frac{1}{2.8 \cdot \exp(\frac{\lfloor \log_2 k \rfloor}{4})} + \xi_1 \cdot \xi_2 \right), \nonumber \\
                   &\text{where}\; \xi_1 \sim \texttt{Bernoulli}(0.5),\, \xi_2 \sim U(0, 0.05).
\end{align}
We repeat the above branch split 6 times,~\ie,~creating a binary tree with depth 6, resulting in $2^0 + \dots + 2^6 = 127$ branches for each class.
For both classes, we have $\mathbf{s}_0 = \mathbf{0}$ and $l_0 = 1.2$.
While class 0 uses $o_0 = 0.25 \pi$ and class 1 uses $o_0 = 1.75 \pi$.

For each branch constructed above, we create 8 Gaussians whose means are uniformly positioned along the branch. 
For each class, for the $i$-th Gaussian from a total of $127 (\text{\#branches}) \times 8 = 1016$ Gaussians, we define the covariance matrix by rotating the base covariance matrix $\texttt{diag}\left( 0.005 \cdot \exp(-\frac{i}{30}), 0.003 \cdot \exp(-\frac{i}{25}) \right)$ using the corresponding branch's orientation.

The final dataset is constructed by sampling points from the Gaussians defined above.
Concretely, for each class, for the $i$-th branch, we sample $\lfloor 1000 \times \exp(- \frac{i}{100})\rfloor$ points from \textit{each} of the 8 Gaussians on the branch.

\subsubsection{Experiment Setup}\label{sec: fractal exp setup}

For the experiments in~\figref{fig: fractal cg}:
To demonstrate that the proposed postprocessing benefits sample generation across various levels of entanglements, we intentionally train a \textit{linear} classifier which struggles to provide a correct signal $p_\theta ( \cond \vert \data_{t})$ in~\equref{eq: guided_diffusion} as we increase the entanglement of both classes  from~\figref{fig: fractal cg 0.0} to~\ref{fig: fractal cg 1.0}.

After training, decision boundaries for classifiers in~\figref{fig: fractal cg 0.0},~\ref{fig: fractal cg 0.5}, and~\ref{fig: fractal cg 1.0} roughly align with the diagonal from top-left to bottom-right.
For each level of entanglement, we generate two sets of samples,
one with guidance scale $w\!=\!1$ and the other with a large scale ($w\!=\!5$ or 400) that pushes generations away from the classifier's decision boundary.

Then for each set, we train a corresponding rectified flow postprocessing model. 
The postprocessing step always improves the generated samples to match the real data, especially on boundaries between the two classes, verifying our classifier-centric understandings as discussed in~\secref{sec: postprocess}:
\begin{itemize}
    \item For scenarios where scale equals 1 in~\figref{fig: fractal cg}, before postprocessing, a large portion of the generations fall into the incorrect category as the classifier signal is not strong enough. With the proposed postprocessing step, we observe correct generations aligned with the conditioning information. 
    \item For large scales, though the generations are generally correct as the signal from the classifier starts to dominate the generation process, there are still outliers as highlighted.
The proposed postprocessing significantly corrects these low-quality generations while not altering already-high-quality generations.
\end{itemize}

\subsubsection{Models}

For the 2D fractal data, both our denoising diffusion model for \cgguide~and \cfgguide~as well as the postprocessing model are based on the structure illustrated in~\figref{fig: mlp point diffusion}.
In all our 2D fractal experiments we use a \textit{linear} classifier for \cgguide, similar to the one discussed in~\secref{sec: model 1d gaussian}.

We train  1) the classifier; 2) unconditional and conditional diffusion models; and 3) the postprocessing model using the optimizer AdamW~\cite{Loshchilov2017DecoupledWD} with a batch size of 4096 points and a learning rate of $10^{-4}$.
The training iterations for the classifier, the denoising diffusion models, and the postprocessing models are 30k, 100k, and 100k.
We use the adaptive solver Dopri5 for the postprocessing model when conducting generations.
We run each evaluation once.

\subsection{Experiments on MNIST}\label{supp sec: mnist exp detail}

\subsubsection{Experiment Setup}\label{supp sec: mnist exp setup}

For~\figref{fig: mnist cg}, we train a classifier (not the one for \cgguide) that achieves almost perfect accuracy on the validation split.
A principal component analysis (PCA) is built on MNIST training data's features extracted from the classifier. With the fitted PCA, we transform 6k (600 for each digit) generations accordingly and visualize the first two components.

\subsubsection{Models}\label{supp sec: impl mnist}

In all our MNIST experiments we use a \textit{linear} classifier for \cgguide, similar to the one discussed in~\secref{sec: model 1d gaussian}. 
The input is a flattened image concatenated with the diffusion step.
Our denoising diffusion models and the postprocessing model for \cgguide~and \cfgguide~are identical to the ones used by~\citet{dhariwal2021diffusion}.\footnote{\label{footnote:guided unet}\url{https://github.com/openai/guided-diffusion/blob/22e0df818350/guided_diffusion/unet.py\#L396}}
We use the following configuration:
\begin{itemize}
    \item \texttt{model\_channels} = 128;
    \item \texttt{num\_res\_blocks} = 2;
    \item \texttt{attention\_resolutions} = (),~\ie,~empty;
    \item \texttt{channel\_mult} = (1, 2).
\end{itemize}

We train  1) the classifier; 2) unconditional and conditional diffusion models; and 3) the postprocessing models using the optimizer AdamW~\cite{Loshchilov2017DecoupledWD} with a batch size of 512 images and a learning rate of $3\cdot10^{-4}$.
The training iterations for the classifier, the denoising diffusion models, and the postprocessing models are 100k, 2k, and 10k.
We use the adaptive solver Dopri5 for the postprocessing model when conducting generations.
We run each evaluation once.

\subsection{Experiments on CIFAR-10}\label{supp sec: cifar10 exp detail}

\subsubsection{Models}\label{supp sec: impl cifar10}

\noindent\textbf{In case of an EDM base model}, our postprocessing model for \cfgguide~used in~\secref{sec: cifar10} is also identical to the ones used by~\citet{dhariwal2021diffusion}.\footref{footnote:guided unet}
We use the following configuration:
\begin{itemize}
    \item \texttt{model\_channels} = 128;
    \item \texttt{num\_res\_blocks} = 4;  %
    \item \texttt{attention\_resolutions} = (2,);
    \item \texttt{channel\_mult} = (1, 2, 2, 2);
    \item \texttt{num\_heads} = 4;
    \item \texttt{num\_head\_channels} = 64;
    \item \texttt{dropout} = 0.1.
\end{itemize}

Following the setup in~\secref{supp sec: nn choice}, we compose a training set of 50k generations sampled with a guidance scale of 1.0, 2.0, and 3.0. We train the postprocessing model using the  Adam~\cite{Kingma2014AdamAM} optimizer with a batch size of 512 images and a learning rate of $5\cdot10^{-4}$ for 400k iterations.
We store the exponential moving average of the model weights with a decay rate of 0.9999.
We maximize the alignment between the capacities of our postprocessing model and the base EDM model while working within our limited computing resources.
Our postprocessing model's number of parameters is 55.97 M, roughly matching the base EDM model's 55.74 M parameters.
We use the adaptive solver Dopri5 for the postprocessing model when conducting generations.
We run each evaluation once.

\begin{figure*}[!t]
    \vspace{0pt} 
    \centering
    \begin{minipage}[t]{\textwidth}
        \begin{subfigure}[t]{0.49\textwidth}
            \centering
            \adjincludegraphics[width=\textwidth,trim={{0.0\width} 0 {0\width} 0},clip]{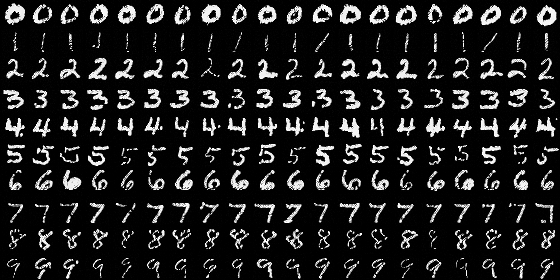}
            \captionsetup{width=\textwidth}
            \caption{
              \textbf{Generations before postprocessing step.}
            }
            \label{fig: mnist cfg before}
        \end{subfigure}
        \hfill
        \begin{subfigure}[t]{0.49\textwidth}
            \centering
            \adjincludegraphics[width=\textwidth,trim={{0.0\width} 0 {0\width} 0},clip]{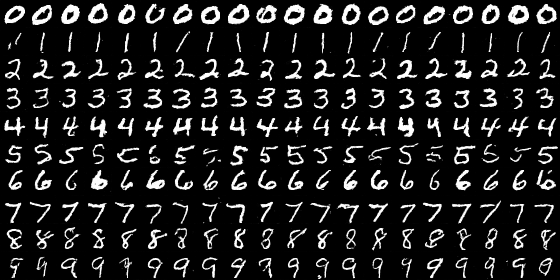}
            \captionsetup{width=\textwidth}
            \caption{
              \textbf{Generations after postprocessing step.}
            }
            \label{fig: mnist cfg after}
        \end{subfigure}
    \end{minipage}%
    \caption{
      \textbf{\Cfgguide~(scale 10.0) with flow-matching based postprocessing (\secref{sec: postprocess}) on MNIST.}
      As expected, a large guidance scale deteriorates  generation quality and  digits are blurry (\figref{fig: mnist cfg before}).
      Our postprocessing  recovers the fidelity as shown in~\figref{fig: mnist cfg after}, verifying our classifier-centric understanding as discussed in~\secref{sec: postprocess}.
    }
    \label{fig: mnist cfg}
\end{figure*}

\begin{figure}[!t]
    \centering
    \includegraphics[width=0.6\columnwidth]{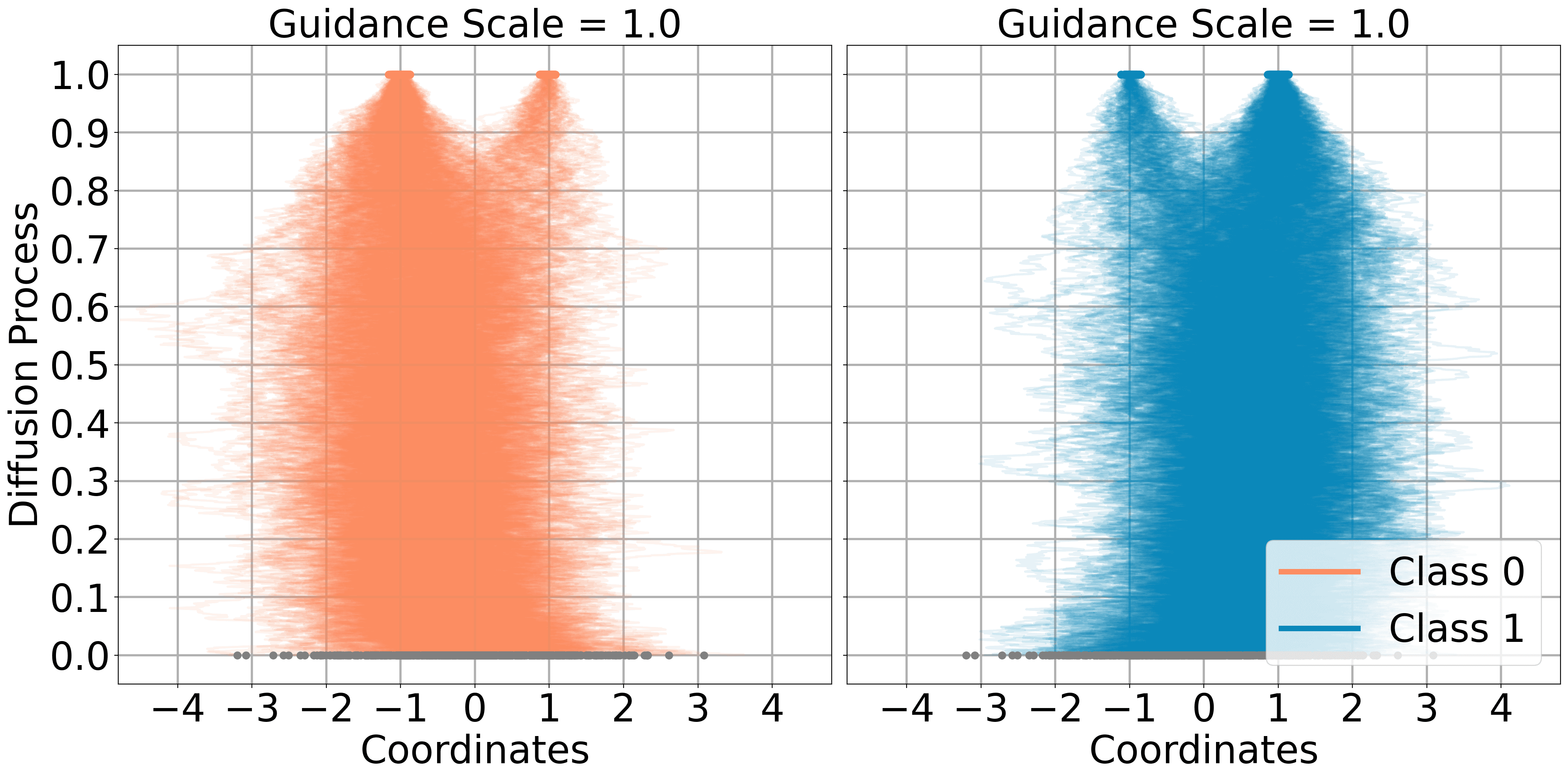}
    \caption{
      \textbf{\Cgguide~with a linear classifier (\figref{fig: gaussian traj cg classifier linear}) per-class visualization.}
    }
    \label{fig: gaussian traj cg classifier linear per class}
\end{figure}

\noindent\textbf{In case of a flow matching base model}, our rectified flow models and the postprocessing model for \cfgguide~used in~\secref{supp sec: nn choice} are identical to the ones used by~\citet{dhariwal2021diffusion}.\footref{footnote:guided unet}
We use the following configuration:
\begin{itemize}
    \item \texttt{model\_channels} = 128;
    \item \texttt{num\_res\_blocks} = 2;
    \item \texttt{attention\_resolutions} = (2,);
    \item \texttt{channel\_mult} = (1, 2, 2, 2);
    \item \texttt{num\_heads} = 4;
    \item \texttt{num\_head\_channels} = 64;
    \item \texttt{dropout} = 0.1.
\end{itemize}

We train both 1) the conditional rectified flow model; and 2) the postprocessing model using the Adam~\cite{Kingma2014AdamAM} optimizer with a batch size of 128 images and a learning rate of $2\cdot10^{-4}$ for 400k iterations.
For the postprocessing model, following the setup in~\secref{supp sec: nn choice}, we compose a training set of 50k generations sampled with a guidance scale of 1.0, 2.0, and 3.0. 
We store the exponential moving average of the model weights with a decay rate of 0.9999.
We run each evaluation once.

\section{Broader Impacts}\label{supp sec: broader impact}

This paper provides an intuitive understanding of the~\cgguide~and~\cfgguide, which could be utilized to improve the generation qualities of diffusion-based generative models.
Similar to many generative modeling techniques, our analysis may be used to recreate license-protected data.

\begin{figure*}[!ht]
    \vspace{0pt} 
    \centering
    \begin{minipage}[t]{\textwidth}
        \begin{subfigure}[t]{\textwidth}
            \centering
            \adjincludegraphics[width=\textwidth,trim={{0.0\width} 0 {0\width} 0},clip]{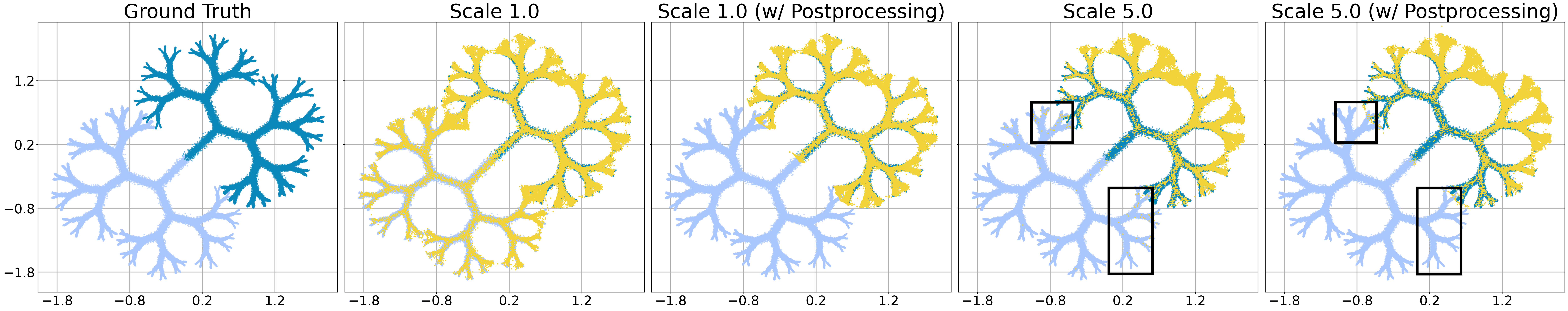}
            \captionsetup{width=\textwidth}
            \caption{
              \textbf{Low entanglement.}
              The $3^\text{rd}$ (and $5^\text{th}$) plot show generated samples after applying postprocessing  on generations from the $2^\text{nd}$ (and $4^\text{th}$) plot.
            }
            \label{fig: fractal cg 0.0 class 0}
        \end{subfigure}
    \end{minipage}%
    \hfill
    \begin{minipage}[t]{\textwidth}
        \begin{subfigure}[t]{\textwidth}
            \centering
            \adjincludegraphics[width=\textwidth,trim={{0.0\width} 0 {0\width} 0},clip]{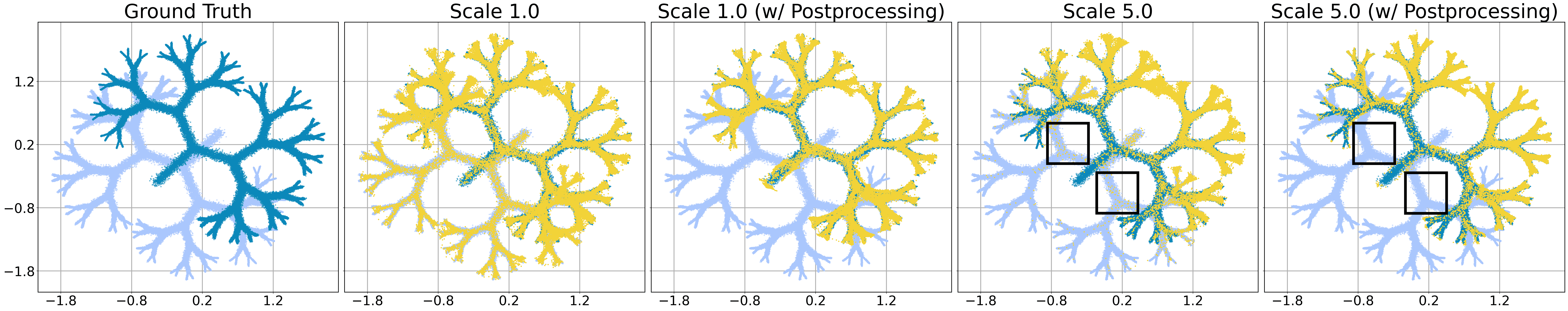}
            \captionsetup{width=\textwidth}
            \caption{
              \textbf{Medium entanglement.}
              The $3^\text{rd}$ (and $5^\text{th}$) plot show generated samples after applying postprocessing  on generations from the $2^\text{nd}$ (and $4^\text{th}$) plot.
            }
            \label{fig: fractal cg 0.5 class 0}
        \end{subfigure}
    \end{minipage}%
    \hfill
    \begin{minipage}[t]{\textwidth}
        \begin{subfigure}[t]{\textwidth}
            \centering
            \adjincludegraphics[width=\textwidth,trim={{0.0\width} 0 {0\width} 0},clip]{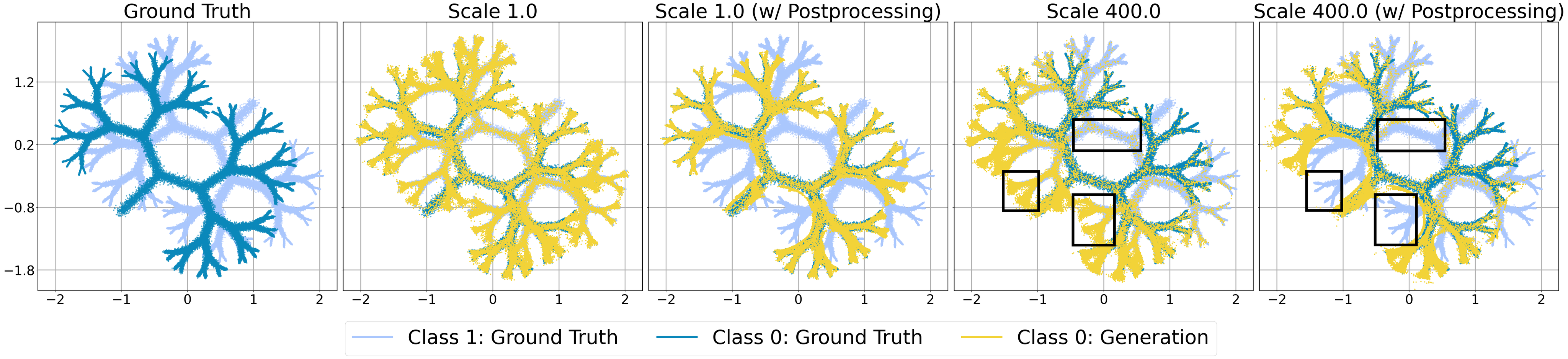}
            \captionsetup{width=\textwidth}
            \caption{
              \textbf{High entanglement.}
              The $3^\text{rd}$ (and $5^\text{th}$) plot show generated samples after applying postprocessing  on generations from the $2^\text{nd}$ (and $4^\text{th}$) plot.
            }
            \label{fig: fractal cg 1.0 class 0}
        \end{subfigure}
    \end{minipage}%
    \caption{
      \textbf{\Cgguide~with flow-matching based postprocessing (\secref{sec: postprocess}) on 2D fractal data.}
      This figure is complementary to~\figref{fig: fractal cg}, showing generations for the other class that are not presented in~\figref{fig: fractal cg}.
      The postprocessing step always improves the generated samples to match the real data, especially on boundaries between the two classes.
      For scenarios where scale equals 1, before postprocessing, a large portion of the generations fall into the incorrect category as the classifier signal is not strong enough. With the proposed postprocessing step, we observe correct generations aligned with the conditioning information. 
      For large scales, though the generations are generally correct as the signal from the classifier starts to dominate the generation process, there are still outliers as highlighted.
      The proposed postprocessing significantly corrects these low-quality generations while not altering already-high-quality generations.
    }
    \label{fig: fractal cg class 0}
\end{figure*}

\section{More Visualizations}\label{sec: more vis}

\figref{fig: mnist cfg} visualizes the effect of our postprocessing on the MNIST dataset.

\figref{fig: gaussian traj cg classifier linear per class} visualizes the per-class denoising diffusion trajectories for the $2^\text{nd}$ column in~\figref{fig: gaussian traj cg classifier linear}. This shows that both classes have a nontrivial portion of incorrect generations.

In~\figref{fig: fractal cg class 0}, we show results the 2D fractal results for the class which was not presented in~\figref{fig: fractal cg}.
As shown in the main paper, our proposed postprocessing step  improves the generation quality around the decision boundaries.

\end{document}